%% file: main.tex
\documentclass{article}

\usepackage{microtype}
\usepackage{graphicx}
\usepackage{booktabs} 

\usepackage{hyperref}
\usepackage{wrapfig}
\usepackage{microtype}      
\usepackage{nicefrac}       
\usepackage{amsfonts}       
\usepackage{booktabs}       
\usepackage{url}            
\usepackage[utf8]{inputenc} 
\usepackage[T1]{fontenc}    

\usepackage[ruled,vlined]{algorithm2e}
\usepackage{algorithmic}

\usepackage[preprint]{icml2024}

\usepackage{amsmath}
\usepackage{amssymb}
\usepackage{mathtools}
\usepackage{amsthm}
\usepackage{pifont}
\usepackage{bm}
\usepackage{multirow}
\usepackage{subcaption}
\usepackage{caption}
\usepackage{xcolor}
\usepackage{colortbl}

\usepackage[capitalize,noabbrev]{cleveref}

\theoremstyle{plain}
\newtheorem{theorem}{Theorem}[section]
\newtheorem{proposition}[theorem]{Proposition}
\newtheorem{lemma}[theorem]{Lemma}
\newtheorem{corollary}[theorem]{Corollary}
\theoremstyle{definition}
\newtheorem{definition}[theorem]{Definition}
\newtheorem{assumption}[theorem]{Assumption}
\theoremstyle{remark}
\newtheorem{remark}[theorem]{Remark}

\newcommand{\Autoref}[1]{%
  \begingroup%
  \def\algorithmautorefname{Algorithm}%
  \def\chapterautorefname{Chapter}%
  \def\sectionautorefname{Section}%
  \def\subsectionautorefname{Subsection}%
  \autoref{#1}%
  \endgroup%
}

\usepackage[textsize=tiny]{todonotes}

\icmltitlerunning{Revisiting Early-Learning Regularization: When Federated Learning Meets Noisy Labels}

\begin{document}

\twocolumn[
\icmltitle{Revisiting Early-Learning Regularization: \\ When Federated Learning Meets Noisy Labels}

\icmlsetsymbol{equal}{*}

\begin{icmlauthorlist}
\icmlauthor{Taehyeon Kim}{yyy}
\icmlauthor{Donggyu Kim}{comp}
\icmlauthor{Se-Young Yun}{yyy}
\end{icmlauthorlist}

\icmlaffiliation{yyy}{KAIST AI} 
\icmlaffiliation{comp}{Medipixel, work done while at KAIST AI}
%
\icmlcorrespondingauthor{Se-Young Yun}{yunseyoung@kaist.ac.kr}

\icmlkeywords{Machine Learning, ICML}

\vskip 0.3in
]



\printAffiliationsAndNotice{}  

\begin{abstract}
In the evolving landscape of federated learning (FL), addressing label noise presents unique challenges due to the decentralized and diverse nature of data collection across clients. Traditional centralized learning approaches to mitigate label noise are constrained in FL by privacy concerns and the heterogeneity of client data. This paper revisits early-learning regularization, introducing an innovative strategy, Federated Label-mixture Regularization (FLR). FLR adeptly adapts to FL's complexities by generating new pseudo labels, blending local and global model predictions. This method not only enhances the accuracy of the global model in both i.i.d. and non-i.i.d. settings but also effectively counters the memorization of noisy labels. Demonstrating compatibility with existing label noise and FL techniques, FLR paves the way for improved generalization in FL environments fraught with label inaccuracies.
\end{abstract}

\input{math_commands}

\input{section/1_intro}
\input{section/2_method}

\input{section/3_exp}

\input{section/4_related_work}
\input{section/5_conclusion}




\newpage

\paragraph{Broader Impact} This paper presents work whose goal is to advance the field of Machine Learning. There are many potential societal consequences of our work, none which we feel must be specifically highlighted in the body. Our opinions on the broader impact (e.g., ehics statement, limitations, and future directions) are discussed in the Appendix.

\bibliography{ref}
\bibliographystyle{icml2024}

\newpage
\appendix
\onecolumn
\input{appendix/app1}
\input{appendix/app2}

\end{document}

%% file: math_commands.tex


\newcommand{\figleft}{{\em (Left)}}
\newcommand{\figcenter}{{\em (Center)}}
\newcommand{\figright}{{\em (Right)}}
\newcommand{\figtop}{{\em (Top)}}
\newcommand{\figbottom}{{\em (Bottom)}}
\newcommand{\captiona}{{\em (a)}}
\newcommand{\captionb}{{\em (b)}}
\newcommand{\captionc}{{\em (c)}}
\newcommand{\captiond}{{\em (d)}}

\newcommand{\newterm}[1]{{\bf #1}}

\def\figref#1{figure~\ref{#1}}
\def\Figref#1{Figure~\ref{#1}}
\def\twofigref#1#2{figures \ref{#1} and \ref{#2}}
\def\quadfigref#1#2#3#4{figures \ref{#1}, \ref{#2}, \ref{#3} and \ref{#4}}
\def\secref#1{section~\ref{#1}}
\def\Secref#1{Section~\ref{#1}}
\def\twosecrefs#1#2{sections \ref{#1} and \ref{#2}}
\def\secrefs#1#2#3{sections \ref{#1}, \ref{#2} and \ref{#3}}
\def\eqref#1{equation~\ref{#1}}
\def\Eqref#1{Equation~\ref{#1}}
\def\plaineqref#1{\ref{#1}}
\def\chapref#1{chapter~\ref{#1}}
\def\Chapref#1{Chapter~\ref{#1}}
\def\rangechapref#1#2{chapters\ref{#1}--\ref{#2}}
\def\algref#1{algorithm~\ref{#1}}
\def\Algref#1{Algorithm~\ref{#1}}
\def\twoalgref#1#2{algorithms \ref{#1} and \ref{#2}}
\def\Twoalgref#1#2{Algorithms \ref{#1} and \ref{#2}}
\def\partref#1{part~\ref{#1}}
\def\Partref#1{Part~\ref{#1}}
\def\twopartref#1#2{parts \ref{#1} and \ref{#2}}

\def\ceil#1{\lceil #1 \rceil}
\def\floor#1{\lfloor #1 \rfloor}
\def\1{\bm{1}}
\newcommand{\train}{\mathcal{D}}
\newcommand{\valid}{\mathcal{D_{\mathrm{valid}}}}
\newcommand{\test}{\mathcal{D_{\mathrm{test}}}}

\def\eps{{\epsilon}}

\def\reta{{\textnormal{$\eta$}}}
\def\ra{{\textnormal{a}}}
\def\rb{{\textnormal{b}}}
\def\rc{{\textnormal{c}}}
\def\rd{{\textnormal{d}}}
\def\re{{\textnormal{e}}}
\def\rf{{\textnormal{f}}}
\def\rg{{\textnormal{g}}}
\def\rh{{\textnormal{h}}}
\def\ri{{\textnormal{i}}}
\def\rj{{\textnormal{j}}}
\def\rk{{\textnormal{k}}}
\def\rl{{\textnormal{l}}}
\def\rn{{\textnormal{n}}}
\def\ro{{\textnormal{o}}}
\def\rp{{\textnormal{p}}}
\def\rq{{\textnormal{q}}}
\def\rr{{\textnormal{r}}}
\def\rs{{\textnormal{s}}}
\def\rt{{\textnormal{t}}}
\def\ru{{\textnormal{u}}}
\def\rv{{\textnormal{v}}}
\def\rw{{\textnormal{w}}}
\def\rx{{\textnormal{x}}}
\def\ry{{\textnormal{y}}}
\def\rz{{\textnormal{z}}}

\def\rvepsilon{{\mathbf{\epsilon}}}
\def\rvtheta{{\mathbf{\theta}}}
\def\rva{{\mathbf{a}}}
\def\rvb{{\mathbf{b}}}
\def\rvc{{\mathbf{c}}}
\def\rvd{{\mathbf{d}}}
\def\rve{{\mathbf{e}}}
\def\rvf{{\mathbf{f}}}
\def\rvg{{\mathbf{g}}}
\def\rvh{{\mathbf{h}}}
\def\rvu{{\mathbf{i}}}
\def\rvj{{\mathbf{j}}}
\def\rvk{{\mathbf{k}}}
\def\rvl{{\mathbf{l}}}
\def\rvm{{\mathbf{m}}}
\def\rvn{{\mathbf{n}}}
\def\rvo{{\mathbf{o}}}
\def\rvp{{\mathbf{p}}}
\def\rvq{{\mathbf{q}}}
\def\rvr{{\mathbf{r}}}
\def\rvs{{\mathbf{s}}}
\def\rvt{{\mathbf{t}}}
\def\rvu{{\mathbf{u}}}
\def\rvv{{\mathbf{v}}}
\def\rvw{{\mathbf{w}}}
\def\rvx{{\mathbf{x}}}
\def\rvy{{\mathbf{y}}}
\def\rvz{{\mathbf{z}}}

\def\erva{{\textnormal{a}}}
\def\ervb{{\textnormal{b}}}
\def\ervc{{\textnormal{c}}}
\def\ervd{{\textnormal{d}}}
\def\erve{{\textnormal{e}}}
\def\ervf{{\textnormal{f}}}
\def\ervg{{\textnormal{g}}}
\def\ervh{{\textnormal{h}}}
\def\ervi{{\textnormal{i}}}
\def\ervj{{\textnormal{j}}}
\def\ervk{{\textnormal{k}}}
\def\ervl{{\textnormal{l}}}
\def\ervm{{\textnormal{m}}}
\def\ervn{{\textnormal{n}}}
\def\ervo{{\textnormal{o}}}
\def\ervp{{\textnormal{p}}}
\def\ervq{{\textnormal{q}}}
\def\ervr{{\textnormal{r}}}
\def\ervs{{\textnormal{s}}}
\def\ervt{{\textnormal{t}}}
\def\ervu{{\textnormal{u}}}
\def\ervv{{\textnormal{v}}}
\def\ervw{{\textnormal{w}}}
\def\ervx{{\textnormal{x}}}
\def\ervy{{\textnormal{y}}}
\def\ervz{{\textnormal{z}}}

\def\rmA{{\mathbf{A}}}
\def\rmB{{\mathbf{B}}}
\def\rmC{{\mathbf{C}}}
\def\rmD{{\mathbf{D}}}
\def\rmE{{\mathbf{E}}}
\def\rmF{{\mathbf{F}}}
\def\rmG{{\mathbf{G}}}
\def\rmH{{\mathbf{H}}}
\def\rmI{{\mathbf{I}}}
\def\rmJ{{\mathbf{J}}}
\def\rmK{{\mathbf{K}}}
\def\rmL{{\mathbf{L}}}
\def\rmM{{\mathbf{M}}}
\def\rmN{{\mathbf{N}}}
\def\rmO{{\mathbf{O}}}
\def\rmP{{\mathbf{P}}}
\def\rmQ{{\mathbf{Q}}}
\def\rmR{{\mathbf{R}}}
\def\rmS{{\mathbf{S}}}
\def\rmT{{\mathbf{T}}}
\def\rmU{{\mathbf{U}}}
\def\rmV{{\mathbf{V}}}
\def\rmW{{\mathbf{W}}}
\def\rmX{{\mathbf{X}}}
\def\rmY{{\mathbf{Y}}}
\def\rmZ{{\mathbf{Z}}}

\def\ermA{{\textnormal{A}}}
\def\ermB{{\textnormal{B}}}
\def\ermC{{\textnormal{C}}}
\def\ermD{{\textnormal{D}}}
\def\ermE{{\textnormal{E}}}
\def\ermF{{\textnormal{F}}}
\def\ermG{{\textnormal{G}}}
\def\ermH{{\textnormal{H}}}
\def\ermI{{\textnormal{I}}}
\def\ermJ{{\textnormal{J}}}
\def\ermK{{\textnormal{K}}}
\def\ermL{{\textnormal{L}}}
\def\ermM{{\textnormal{M}}}
\def\ermN{{\textnormal{N}}}
\def\ermO{{\textnormal{O}}}
\def\ermP{{\textnormal{P}}}
\def\ermQ{{\textnormal{Q}}}
\def\ermR{{\textnormal{R}}}
\def\ermS{{\textnormal{S}}}
\def\ermT{{\textnormal{T}}}
\def\ermU{{\textnormal{U}}}
\def\ermV{{\textnormal{V}}}
\def\ermW{{\textnormal{W}}}
\def\ermX{{\textnormal{X}}}
\def\ermY{{\textnormal{Y}}}
\def\ermZ{{\textnormal{Z}}}

\def\vzero{{\bm{0}}}
\def\vone{{\bm{1}}}
\def\vmu{{\bm{\mu}}}
\def\vtheta{{\bm{\theta}}}
\def\va{{\bm{a}}}
\def\vb{{\bm{b}}}
\def\vc{{\bm{c}}}
\def\vd{{\bm{d}}}
\def\ve{{\bm{e}}}
\def\vf{{\bm{f}}}
\def\vg{{\bm{g}}}
\def\vh{{\bm{h}}}
\def\vi{{\bm{i}}}
\def\vj{{\bm{j}}}
\def\vk{{\bm{k}}}
\def\vl{{\bm{l}}}
\def\vm{{\bm{m}}}
\def\vn{{\bm{n}}}
\def\vo{{\bm{o}}}
\def\vp{{\bm{p}}}
\def\vq{{\bm{q}}}
\def\vr{{\bm{r}}}
\def\vs{{\bm{s}}}
\def\vt{{\bm{t}}}
\def\vu{{\bm{u}}}
\def\vv{{\bm{v}}}
\def\vw{{\bm{w}}}
\def\vx{{\bm{x}}}
\def\vy{{\bm{y}}}
\def\vz{{\bm{z}}}

\def\evalpha{{\alpha}}
\def\evbeta{{\beta}}
\def\evepsilon{{\epsilon}}
\def\evlambda{{\lambda}}
\def\evomega{{\omega}}
\def\evmu{{\mu}}
\def\evpsi{{\psi}}
\def\evsigma{{\sigma}}
\def\evtheta{{\theta}}
\def\eva{{a}}
\def\evb{{b}}
\def\evc{{c}}
\def\evd{{d}}
\def\eve{{e}}
\def\evf{{f}}
\def\evg{{g}}
\def\evh{{h}}
\def\evi{{i}}
\def\evj{{j}}
\def\evk{{k}}
\def\evl{{l}}
\def\evm{{m}}
\def\evn{{n}}
\def\evo{{o}}
\def\evp{{p}}
\def\evq{{q}}
\def\evr{{r}}
\def\evs{{s}}
\def\evt{{t}}
\def\evu{{u}}
\def\evv{{v}}
\def\evw{{w}}
\def\evx{{x}}
\def\evy{{y}}
\def\evz{{z}}

\def\mA{{\bm{A}}}
\def\mB{{\bm{B}}}
\def\mC{{\bm{C}}}
\def\mD{{\bm{D}}}
\def\mE{{\bm{E}}}
\def\mF{{\bm{F}}}
\def\mG{{\bm{G}}}
\def\mH{{\bm{H}}}
\def\mI{{\bm{I}}}
\def\mJ{{\bm{J}}}
\def\mK{{\bm{K}}}
\def\mL{{\bm{L}}}
\def\mM{{\bm{M}}}
\def\mN{{\bm{N}}}
\def\mO{{\bm{O}}}
\def\mP{{\bm{P}}}
\def\mQ{{\bm{Q}}}
\def\mR{{\bm{R}}}
\def\mS{{\bm{S}}}
\def\mT{{\bm{T}}}
\def\mU{{\bm{U}}}
\def\mV{{\bm{V}}}
\def\mW{{\bm{W}}}
\def\mX{{\bm{X}}}
\def\mY{{\bm{Y}}}
\def\mZ{{\bm{Z}}}
\def\mBeta{{\bm{\beta}}}
\def\mPhi{{\bm{\Phi}}}
\def\mLambda{{\bm{\Lambda}}}
\def\mSigma{{\bm{\Sigma}}}

\newcommand{\tens}[1]{\bm{\mathsfit{#1}}}
\def\tA{{\tens{A}}}
\def\tB{{\tens{B}}}
\def\tC{{\tens{C}}}
\def\tD{{\tens{D}}}
\def\tE{{\tens{E}}}
\def\tF{{\tens{F}}}
\def\tG{{\tens{G}}}
\def\tH{{\tens{H}}}
\def\tI{{\tens{I}}}
\def\tJ{{\tens{J}}}
\def\tK{{\tens{K}}}
\def\tL{{\tens{L}}}
\def\tM{{\tens{M}}}
\def\tN{{\tens{N}}}
\def\tO{{\tens{O}}}
\def\tP{{\tens{P}}}
\def\tQ{{\tens{Q}}}
\def\tR{{\tens{R}}}
\def\tS{{\tens{S}}}
\def\tT{{\tens{T}}}
\def\tU{{\tens{U}}}
\def\tV{{\tens{V}}}
\def\tW{{\tens{W}}}
\def\tX{{\tens{X}}}
\def\tY{{\tens{Y}}}
\def\tZ{{\tens{Z}}}

\def\gA{{\mathcal{A}}}
\def\gB{{\mathcal{B}}}
\def\gC{{\mathcal{C}}}
\def\gD{{\mathcal{D}}}
\def\gE{{\mathcal{E}}}
\def\gF{{\mathcal{F}}}
\def\gG{{\mathcal{G}}}
\def\gH{{\mathcal{H}}}
\def\gI{{\mathcal{I}}}
\def\gJ{{\mathcal{J}}}
\def\gK{{\mathcal{K}}}
\def\gL{{\mathcal{L}}}
\def\gM{{\mathcal{M}}}
\def\gN{{\mathcal{N}}}
\def\gO{{\mathcal{O}}}
\def\gP{{\mathcal{P}}}
\def\gQ{{\mathcal{Q}}}
\def\gR{{\mathcal{R}}}
\def\gS{{\mathcal{S}}}
\def\gT{{\mathcal{T}}}
\def\gU{{\mathcal{U}}}
\def\gV{{\mathcal{V}}}
\def\gW{{\mathcal{W}}}
\def\gX{{\mathcal{X}}}
\def\gY{{\mathcal{Y}}}
\def\gZ{{\mathcal{Z}}}

\def\sA{{\mathbb{A}}}
\def\sB{{\mathbb{B}}}
\def\sC{{\mathbb{C}}}
\def\sD{{\mathbb{D}}}
\def\sF{{\mathbb{F}}}
\def\sG{{\mathbb{G}}}
\def\sH{{\mathbb{H}}}
\def\sI{{\mathbb{I}}}
\def\sJ{{\mathbb{J}}}
\def\sK{{\mathbb{K}}}
\def\sL{{\mathbb{L}}}
\def\sM{{\mathbb{M}}}
\def\sN{{\mathbb{N}}}
\def\sO{{\mathbb{O}}}
\def\sP{{\mathbb{P}}}
\def\sQ{{\mathbb{Q}}}
\def\sR{{\mathbb{R}}}
\def\sS{{\mathbb{S}}}
\def\sT{{\mathbb{T}}}
\def\sU{{\mathbb{U}}}
\def\sV{{\mathbb{V}}}
\def\sW{{\mathbb{W}}}
\def\sX{{\mathbb{X}}}
\def\sY{{\mathbb{Y}}}
\def\sZ{{\mathbb{Z}}}

\def\emLambda{{\Lambda}}
\def\emA{{A}}
\def\emB{{B}}
\def\emC{{C}}
\def\emD{{D}}
\def\emE{{E}}
\def\emF{{F}}
\def\emG{{G}}
\def\emH{{H}}
\def\emI{{I}}
\def\emJ{{J}}
\def\emK{{K}}
\def\emL{{L}}
\def\emM{{M}}
\def\emN{{N}}
\def\emO{{O}}
\def\emP{{P}}
\def\emQ{{Q}}
\def\emR{{R}}
\def\emS{{S}}
\def\emT{{T}}
\def\emU{{U}}
\def\emV{{V}}
\def\emW{{W}}
\def\emX{{X}}
\def\emY{{Y}}
\def\emZ{{Z}}
\def\emSigma{{\Sigma}}

\newcommand{\etens}[1]{\mathsfit{#1}}
\def\etLambda{{\etens{\Lambda}}}
\def\etA{{\etens{A}}}
\def\etB{{\etens{B}}}
\def\etC{{\etens{C}}}
\def\etD{{\etens{D}}}
\def\etE{{\etens{E}}}
\def\etF{{\etens{F}}}
\def\etG{{\etens{G}}}
\def\etH{{\etens{H}}}
\def\etI{{\etens{I}}}
\def\etJ{{\etens{J}}}
\def\etK{{\etens{K}}}
\def\etL{{\etens{L}}}
\def\etM{{\etens{M}}}
\def\etN{{\etens{N}}}
\def\etO{{\etens{O}}}
\def\etP{{\etens{P}}}
\def\etQ{{\etens{Q}}}
\def\etR{{\etens{R}}}
\def\etS{{\etens{S}}}
\def\etT{{\etens{T}}}
\def\etU{{\etens{U}}}
\def\etV{{\etens{V}}}
\def\etW{{\etens{W}}}
\def\etX{{\etens{X}}}
\def\etY{{\etens{Y}}}
\def\etZ{{\etens{Z}}}

\newcommand{\pdata}{p_{\rm{data}}}
\newcommand{\ptrain}{\hat{p}_{\rm{data}}}
\newcommand{\Ptrain}{\hat{P}_{\rm{data}}}
\newcommand{\pmodel}{p_{\rm{model}}}
\newcommand{\Pmodel}{P_{\rm{model}}}
\newcommand{\ptildemodel}{\tilde{p}_{\rm{model}}}
\newcommand{\pencode}{p_{\rm{encoder}}}
\newcommand{\pdecode}{p_{\rm{decoder}}}
\newcommand{\precons}{p_{\rm{reconstruct}}}

\newcommand{\laplace}{\mathrm{Laplace}} 

\newcommand{\E}{\mathbb{E}}
\newcommand{\Ls}{\mathcal{L}}
\newcommand{\R}{\mathbb{R}}
\newcommand{\emp}{\tilde{p}}
\newcommand{\lr}{\alpha}
\newcommand{\reg}{\lambda}
\newcommand{\rect}{\mathrm{rectifier}}
\newcommand{\softmax}{\mathrm{softmax}}
\newcommand{\sigmoid}{\sigma}
\newcommand{\softplus}{\zeta}
\newcommand{\KL}{D_{\mathrm{KL}}}
\newcommand{\Var}{\mathrm{Var}}
\newcommand{\standarderror}{\mathrm{SE}}
\newcommand{\Cov}{\mathrm{Cov}}
\newcommand{\normlzero}{L^0}
\newcommand{\normlone}{L^1}
\newcommand{\normltwo}{L^2}
\newcommand{\normlp}{L^p}
\newcommand{\normmax}{L^\infty}

\newcommand{\parents}{Pa} 

\let\ab\allowbreak

%% file: section/1_intro.tex
\section{Introduction}

The advent of large-scale datasets has propelled deep neural networks to remarkable achievements in fields such as computer vision, information retrieval, and language processing\,\cite{yao2021instance}. However, these datasets often contain sensitive personal information, making conventional centralized learning impractical for applications such as person identification, financial services, and healthcare systems. Federated Learning\,(FL) addresses privacy issues by allowing clients (e.g., edge devices) to collaborate with a central server (e.g., a service manager) without sharing their local data. Instead, clients update their local models using their private data, and the central server aggregates these updates to improve the global model. This procedure is repeated until convergence\,(FedAvg)\,\cite{fedavg}. FL enables on-device learning and avoids systematic privacy risks at the data level, making it particularly suitable for the edge computing devices such as phones and tablets.

In FL, multiple clients may have different levels of label noise~\cite{fedcorr} due to various factors, such as annotator skill, client bias, malfunctioning data collectors\,\cite{kim2018deep}, or even malicious tampering with labels\,\cite{focus}. For instance, in the healthcare industry, manually labeling patient records at each hospital is susceptible to corruption due to the complexity of medical terminology and the potential for annotator bias. Additionally, data entry mistakes or misunderstandings by medical professionals may introduce inconsistencies or errors into the original data\,\cite{xu2019l_dmi,karimi2020deep}. 
Training with noisy labels can negatively impact over-parameterized neural networks, as they are prone to memorizing mislabeled instances (a.k.a., memorization). Label noise, combined with FL's heterogeneous nature, can exacerbate this phenomenon. Traditional methods for addressing label noise in Centralized Learning\,(CL) are not directly applicable in the FL setting due to privacy constraints, small client datasets, and difficulties in handling data heterogeneity\,\cite{focus, fedcorr}.

\begin{figure*}[t]
    \centerline{\includegraphics[width=0.7\linewidth]{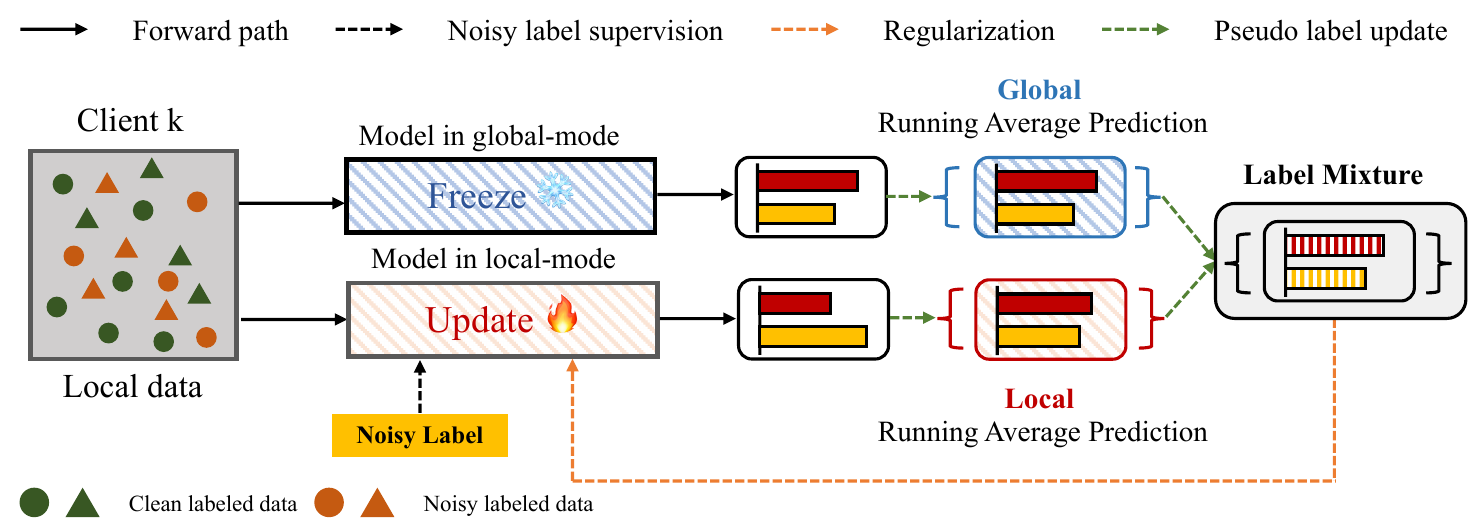}}
    \caption{Overview of our proposed regularization, Federated Label-mixture Regularization (FLR).} \label{fig:overview_method}
    \vspace{-10pt}
\end{figure*}

In this work, we first identifies and addresses the dual nature of memorization in FL – local memorization at the client level and global memorization at the server level (Section~\ref{sec3}). We explore how local memorization, driven by client-specific data and biases, poses a significant risk of overfitting to noisy labels. Conversely, global memorization highlights the challenges faced by the central model in distilling accurate information from diverse, and potentially noisy, client updates. Our analysis reveals the intricate interplay between these two forms of memorization and their cumulative impact on the learning process in FL.

To combat these challenges, we introduce a simple yet efficient novel regularization for dealing with noisy labels in FL, termed as \underline{F}ederated \underline{L}abel-mixture \underline{R}egularization\,(\textbf{FLR}; \autoref{fig:overview_method}), which supervises the noisy instances with pseudo labels generated through a combination of global server's running average predictions and local running average predictions. Our method is grounded in the recognition that memorization in FL manifests distinctively at server and client levels. In both i.i.d. and non-i.i.d. scenarios, the server's and client's temporal ensembling vectors each mitigate memorization issues. However, neither alone fully addresses both local and global memorization. This underscores the strength of our FLR method, which combines these vectors to effectively handle various noise conditions and data heterogeneity.
Through experimental validation, we ascertain that leveraging a mixture of both vectors culminates in a robust label mixture adept at handling various noise conditions and data heterogeneity. Building upon this, we further develop several techniques by balancing the scale between global running average predictions and local running average predictions. In light of the above, our key contributions are as follows:
 
\begin{itemize}
    \item We conduct exploratory experiments to analyze memorization patterns in FL, focusing on both client and server sides, to understand how label noise affects learning (\textbf{\Autoref{sec3}}).

    \item Our novel method, FLR, enhances global model accuracy in FL without extra stages or compromising data privacy, aligning with the decentralized nature of FL\,(\textbf{\Autoref{sec3}}).

    \item 
    We demonstrate that FLR effectively prevents local and global memorization and outperforms other methods on benchmark datasets with various levels of data heterogeneity and label noise\,(\textbf{\Autoref{sec3}}, \textbf{\Autoref{sec4}}).

\end{itemize}

%% file: section/2_method.tex
\section{Preliminaries}\label{problem setup}

The objective of FL is to solve the optimization problem for a distributed collection of heterogeneous data: $\label{eq:FL}\min_\vw\,f(\vw) := \sum_{k\in S}\frac{n_k}{n}F_k(\vw)$
where $S$ is the set of total clients, with each client $k$ has $n_k$ local training data samples, and $n=\sum_{k\in S}n_k$.
The local objective of client $k$ is to minimize $F_k(\vw) = \mathbb{E}_{x_k\sim\mathcal{D}_k} [\ell_k(\vx_k,\vy_k;\vw)]$, where $\ell_k$ is the loss function parameterized by $\vw$ on the local data $(\vx_k,\vy_k)$ from local data distribution $\mathcal{D}_k$. Here, we consider a setting having $N$ clients and dataset with $C$ classes. Each client $k$ has the training set $\{ ({\vx}_{k_i}, {\vy}_{k_i}) \}_{i=1}^{n_k}$, where $\vx_{k_i}$ is the $i$\,th input and $\vy_{k_i}\in\{0,1\}^C$ is the corresponding one-hot label vector; $\vy_{k_i}^{(c)} = 1$ iff $\vx_{k_i}$ belongs to class $c$. The main issue is that it is unknown whether such labels are correctly annotated\,(i.e., clean) or not\,(i.e., noisy).

\paragraph{Data Heterogeneity} 

In order to consider the data heterogeneity in FL, we assume the true data distribution $\mathcal{D}_k$ prior to introducing label noise. In i.i.d. setting, all clients have the same size local training set and an equal number of data samples per class. In contrast, the local data distributions in non-i.i.d. settings are more complex, with varying local training set sizes and imbalanced data samples per class. Following the settings in \citet{fedcorr}, we create a non-i.i.d scenario. We first sample a Bernoulli random variable $\Phi_{kc} \sim \operatorname{Bernoulli}(p)$ for each client $k$ and class $c$, where $\Phi_{kc} = 1$ represents that client $k$ has class $c$ and $\Phi_{kc} = 0$ otherwise. We then distribute the data samples of class $c$ among the clients with $\Phi_{*c} = 1$ using Latent Dirichlet Allocation (LDA), assigning the partition of the class $c$ samples, where $\bm \alpha_c$ is a vector of length $\sum_{k=1}^N \Phi_{kc}$ with positive elements $\alpha_{Dir}$. To ensure that no clients are left without data, we allocate at least one sample of class $c$ to each client $k$ with $\Phi_{kc} = 1$ regardless of $\vq_c$. 

\begin{figure*}[t]
    \centerline{\includegraphics[width=1.0\linewidth]{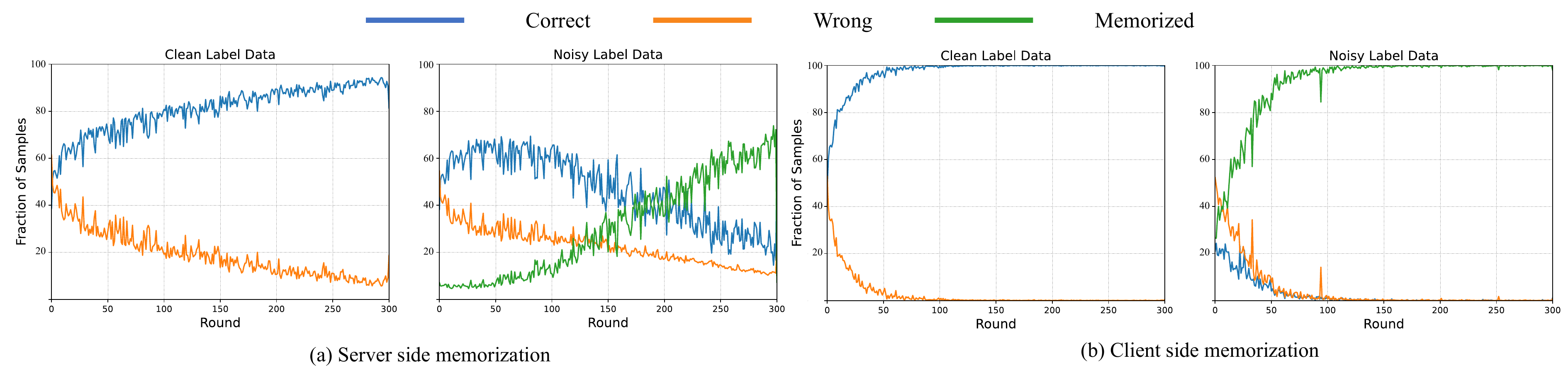}}
    \caption{(a) Server side memorization with $\mathcal{L}_{CE}$, and (b) client side memorization with $\mathcal{L}_{CE}$ on CIFAR-10 of the i.i.d. setting, with symmetric noise of $(\rho,\tau)=(0.8,0.0)$. In (b), the fraction values are calculated by averaging the values contributed solely by participating noisy clients in each round.} \label{fig:general_memorization}
    \vspace{-10pt}
\end{figure*}

\paragraph{Noise Level} 

We synthetically design scenarios with different levels of label noise among clients, following \citet{fedcorr}. We first introduce the parameter $\rho$, which represents the ratio of clients with label noise in their datasets. We then assign the noise rate of each noisy client based on $\tau$, the lower bound of each client's noise rate. We sample the noise rate $r_k$ of noisy client $k$ uniformly at random from the range $\mathcal{U}(\tau, 1)$. Finally, we randomly select $r_k n_k$ data samples and reassign their labels to create noise. It is worth noting that, while the original paper by FedCorr treated $\rho$ and $r_k$ as probabilities (through Bernoulli sampling), our implementation treats them as fixed ratios.

\paragraph{Noise Type}

We examine two forms of label noise: symmetric and asymmetric noise\,\cite{noisetype}. For symmetric noise, we randomly and uniformly choose one class among the $C$ classes and reassign $\vy_{ki}$ as an i.i.d. random one-hot vector. As for asymmetric noise, it mimics human-like errors that happen for similar classes (e.g., cat $\leftrightarrow$ dog, truck $\rightarrow$ automobile). We reassign $\vy_{ki}$ based on the class of $\vx_{ki}$, the initial value of $\vy_{ki}$, for asymmetric noise.

\section{Method}\label{sec3}

\subsection{In the View of Memorization in FL}
In this subsection, we delve into the phenomenon of memorization in FL. Despite numerous studies in CL exploring memorization, the topic remains notably unexplored in the context of FL. To provide a clear understanding, we distinguish between two types of memorization by referring to the definition of memorization in the field of learning with noisy labels in CL\,\cite{elr}:

\begin{itemize}
    \item \textbf{Local memorization (client side)} takes place immediately after a local model update and refers to the accuracy computed on the training dataset using a personalized model that has been updated locally before federated aggregation occurs.
    \item \textbf{Global memorization (server side)} transpires after federated aggregation when the accuracy of the training dataset is calculated using the server model.
\end{itemize}

\autoref{fig:general_memorization} showcases both local and global memorization. In this context, we estimate memorization as instances where the model's predictions coincide with incorrect labels. We observe that the accuracy curves of correct samples in both clean and noisy label plots increase during the early stages. However, in the noisy label plot, the accuracy of correct samples declines after reaching a certain point, while the level of memorization rises. In contrast, the clean label curve remains stable (\autoref{fig:general_memorization} (a)). In both cases, the count of wrong-predicted samples consistently decreases. Alongside examining the server-side, we also probe into the client-side memorization. Owing to the scarce amount of data available to each client, memorization becomes more prominent, regardless of whether the data is clean or noisy (\autoref{fig:general_memorization} (b)). Specifically, the proportion of samples with incorrect labels memorized by the local model increases rapidly as the round progresses, eventually reaching 100\%. This signifies that the local model has memorized all of the noisy labels present in the client's local dataset.

\begin{figure}[ht]
\begin{algorithm}[H]
\DontPrintSemicolon
    \caption{FLR Algorithm}
    \label{al:flr}
    \begin{algorithmic}[1]
    \SetKwInOut{Input}{Input}
    \INPUT: server model $\mathcal{\theta}_{server}$, randomly initialized parameter $\mathcal{\theta}_0$, client $k$'s model $\mathcal{\theta}_{k}$, local training epoch $E$, learning rate $\eta$, client $k$'s dataset $\mathcal{D}^k = \{(\vx_{k_i}, \vy_{k_i}) \}_{i=1}^{n_k}$\\
    \STATE $\mathcal{\theta}_{server} \leftarrow \mathcal{\theta}_{0}$ \tcp*{Initialization} 
    \textcolor{blue}{/* Phase 1: Warmup FedAvg with standard CE loss */}\\
    \FOR{$e \leftarrow 0, \dots , E-1$}
            \STATE $S^e \leftarrow \textsc{SampleClients}$ \\
            \FOR{each client $k \in S^t$ in parallel}
                \STATE $\theta_{k} \leftarrow \theta_{server}$ \\
                    \STATE $\mathcal{\theta}_{k} \leftarrow \mathcal{\theta}_{k} - \eta \, \nabla (- \frac{1}{n_k}\sum_{i=1}^{n_k} \sum_{c=1}^{C} \vy_{k_i}^{(c)} \log \vp_{k_i}^{(c)}))$ \\
            \ENDFOR
            \STATE   $\mathcal{\theta}_{server} \leftarrow \sum_{u \in S^t} \frac{|\mathcal{D}^{u}|}{\sum_{k \in S^t} |\mathcal{D}^{k'}|} \mathcal{\theta}_{k}$ 
    \ENDFOR \\
    \textcolor{blue}{/* Phase 2: FedAvg with FLR loss*/}\\
    \FOR{$e \leftarrow 0, \dots , E-1$}
        \STATE $S^e \leftarrow \textsc{SampleClients}$ \\
        \FOR{each client $u \in S^t$ in parallel}
            \STATE $\theta_{k} \leftarrow \theta_{server}$ \\
                \FOR{$i \leftarrow 1, \dots, n_k$}
                    \STATE $\vs_{k_i} \leftarrow \beta \vs_{k_i} + (1-\beta) \vp_{k_i}^{server}$ \tcp*{Global}
                    \STATE $\vm_{k_i} \leftarrow \gamma\vm_{k_i}+(1-\gamma) \vp_{k_i}$ \tcp*{Local}
                    \STATE $\vt_{k_i} \leftarrow \alpha\vs_{k_i}+(1-\alpha)\vm_{k_i}$ \tcp*{Mixture}
                \ENDFOR
                \STATE $\mathcal{\theta}_{k} \leftarrow \mathcal{\theta}_{k} - \eta \, \nabla (- \frac{1}{n_k}\sum_{i=1}^{n_k} \sum_{c=1}^{C} \vy_{k_i}^{(c)} \log \vp_{k_i}^{(c)} + \frac{\lambda}{n_k}\sum_{i=1}^{n_k} \log\left(1- \langle\vp_{k_i}\cdot \vt_{k_i}\rangle\right))$ \\
            \tcp*{Client-Update with FLR}
        \ENDFOR
            \STATE   $\mathcal{\theta}_{server} \leftarrow \sum_{u \in S^t} \frac{|\mathcal{D}^{u}|}{\sum_{k \in S^t} |\mathcal{D}^{k'}|} \mathcal{\theta}_{k}$ 
        \ENDFOR \\
\end{algorithmic}
\end{algorithm}
\vspace{-20pt}
\end{figure}

\begin{figure*}[t]
    \centerline{\includegraphics[width=1.0\linewidth]{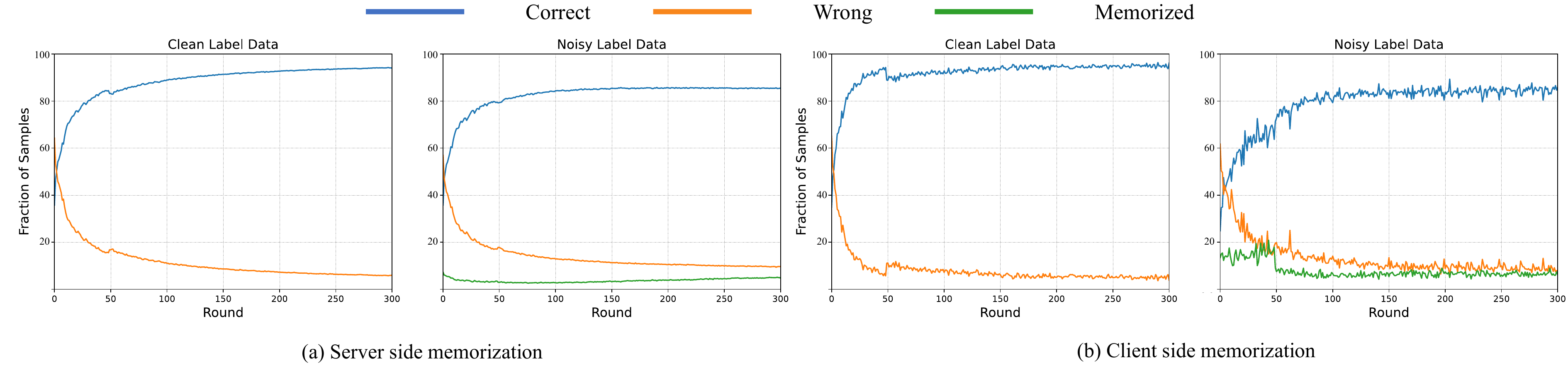}}
    \caption{(a) Server-side memorization with $\mathcal{L}_{FLR}$, and (b) client-side memorization with $\mathcal{L}_{FLR}$ under the same setting used in  \autoref{fig:general_memorization}.} \label{fig:flr_general_memo}
\end{figure*}

Based on this observation, we define memorization from both local and global perspectives, capturing severity and prevalence, respectively. Catastrophic memorization occurs at the client level when a model nearly memorizes its noisy training data, leading to suboptimal generalization and overfitting to incorrect labels. Conversely, common memorization refers to shared memorization across clients from the global perspective, emerging when numerous client models exhibit similar patterns of memorizing noisy labels, indicating a more pervasive issue affecting the overall FL process.

\subsection{Local \& Global Moving Average Label Mixture}
To combat these memorization issues, we introduce a regularization term that penalizes the local prediction based on the difference between the model's output and synthetic pseudo labels:
\vspace{-10pt}

\begin{equation}\label{eq:temporal}
\begin{gathered}
        \mathcal{L}_{FLR}^k (\mathcal{\theta}) \coloneqq \mathcal{L}_{CE}^k (\mathcal{\theta})+ \frac{\lambda}{n_k}\sum_{i=1}^{n_k} \log\left(1- \langle\vp_{k_i}\cdot \vt_{k_i}\rangle\right) \\
        \mathcal{L}_{CE}^k (\mathcal{\theta}) \coloneqq - \frac{1}{n_k}\sum_{i=1}^{n_k} \sum_{c=1}^{C} \vy_{k_i}^{(c)} \log \vp_{k_i}^{(c)}\\
\end{gathered}
\end{equation}
\
\begin{equation*}
\begin{gathered}
\begin{aligned}
    \vs_{k_i} & \leftarrow \beta \vs_{k_i} + (1-\beta) \vp_{k_i}^{server} \vartriangleright	 \text{Global}\\
    \vm_{k_i} & \leftarrow \gamma\vm_{k_i}+(1-\gamma) \vp_{k_i} \quad\, \vartriangleright \text{Local}\\
    \vt_{k_i} & \leftarrow \alpha\vs_{k_i}+(1-\alpha)\vm_{k_i} \quad\, \vartriangleright \text{Mixture}
\end{aligned}
\end{gathered}
\end{equation*}

where $\mathcal{\theta}$ denotes the parameters of the neural network, $\vp_{k_i}$ is the local prediction, $\vt_{k_i}$ is the pseudo label vector for a client $k$'s $i$\,th input, $\vs_{k_i}$ is the global running average prediction, $\vp_{k_i}^{server}$ is the global model's prediction, $\vm_{k_i}$ is a local running average prediction, and $c$ is the index of the vector, which satisfies with $1 \leq c \leq C$. The logarithm component serves as a regularization term that aligns $\vp$ in the same direction as $\vt_{k_i}$. Synthetic pseudo labels $\vt_{k_i}$ are generated by merging a local running average prediction with the server's running average prediction, called Federated Label-mixture Regularization (FLR). For an initial prediction of a client $k$'s $i$\,th input, $\vs_{k_i}$ and $\vm_{k_i}$ are set to be equal to $\vp_{k_i}^{server}$ and $\vp_{k_i}$, respectively. Since the server receives randomly augmented data during each local iteration, $\vp_{k_i}^{server}$ undergoes slight changes. The number of updates to $\vt_{k_i}$ depends on the number of clients participating in each round. If a client does not participate, the most recent version of $\vt_{k_i}$ is stored locally, and this value is used for updates when the client is selected from the server. Algorithm~\ref{al:flr} illustrates the overall procedure of FLR within the general FL framework.

\paragraph{Tackling Confirmation Bias}
The phenomenon of models overfitting to noisy labels, known as memorization, can be viewed as a form of confirmation bias, where incorrect pseudo-labels predicted by the network are erroneously reinforced\,\cite{tarvainen2017mean,arazo2020pseudo}. FLR uses pseudo labels $\vt_{k_i}$ —derived from a blend of local and global running average predictions—to temper this bias. The local predictions capture client-specific noise patterns, while the global predictions reflect the broader data distribution, together creating a comprehensive and resilient learning signal. The balancing parameter $\alpha$ plays a pivotal role, meticulously adjusting the blend of local and global perspectives to ensure optimal noise resistance and overall performance. While local running average prediction alone can be effective in centralized settings, it may exacerbate local overfitting more swiftly in FL contexts due to data heterogeneity and data shortage. Our strategy of blending local and global predictions can thus be particularly advantageous in such situations. More details are in subsection \ref{subsec:implement}.

\paragraph{Dissecting Gradients} The regularization term's logarithm in Eq.~(\ref{eq:temporal}) works to balance out the exponential nature of the softmax function that is present in the probability of $\vp_{k_i}$. Similar to \citet{elr}, the gradient of our method is calculated as follows:

\vspace{-15pt}
\begin{equation}\small \label{gradient}
\begin{gathered}
    \nabla \mathcal{L}_{FLR} (\mathcal{\theta}) = \frac{1}{n} \sum_{i=1}^n \nabla \mathcal{N}_{\vx_{k_i}} (\overbrace{\vp_{k_i} - \vy_{k_i}}^{\text{gradient of } \mathcal{L}_{CE}} + \overbrace{\lambda \vg_{k_i}}^{\text{gradient of the regularization}})
    \\
    \text{where} \,\, \vg_{k_i}^{(c)} \coloneqq \frac{\vp_{k_i}^{(c)}}{1- \langle \vp_{k_i}, \vt_{k_i} \rangle} \sum_{r=1}^C (\vt_{k_i}^{(r)} - \vt_{k_i}^{(c)})\vp_{k_i}^{(r)}
\end{gathered}
\end{equation}

where $\nabla \mathcal{N}_{\vx_{k_i}}$ is the Jacobian matrix of the neural network $\mathcal{N}$ having parameters $\mathcal{\theta}$ for a client $k$'s $i$\,th input with respect to the $\mathcal{\theta}$. Suppose the actual class is $c^*$. In the early training phase, the $c^*$\,th entry of $\vs_{k_i}$ is dominant. Setting $\alpha \geq 0.5$ results in a negative $c^*$\,th entry of $\vg_{k_i}$, since the server memorizes slower than clients, preventing harmful memorization by noisy clients. Moreover, in the case of clean labels, the gradient of $\mathcal{L}_{CE}$ term $\vp_{k_i} - \vy_{k_i}$ decreases quickly due to small dataset size, potentially leading to dominance by wrong labels in the gradient. The term $\vg_{k_i}$ mitigates this by maintaining the magnitude of coefficients on clean label examples. For mislabeled examples, the gradient of $\mathcal{L}_{CE}$ term $\vp_{k_i}^{(r)} - \vy_{k_i}^{(r)}$ is positive, and adding the negative term $\vg_{k_i}^{(r)}$ reduces the coefficients on these examples, minimizing their gradient impact. Beyond a convergence point, the server might not curtail local memorization due to overfitting. The term $\vm_{k_i}$ can be added at this stage to counteract local memorization with local running average predictions.

\begin{table*}[t]
\centering
\caption{Server side memorization and client side memorization at last round on CIFAR-10 of non-i.i.d. setting $(\alpha_{dir} = 1.0)$ with symmetric noise $(\rho, \tau) = (1.0, 0.0)$ and i.i.d. setting with symmetric noise $(\rho, \tau) = (0.8, 0.0)$. Vanilla denotes no use of the regularization in Eq.~(\ref{eq:temporal}).}
\vspace{-5pt}
\resizebox{\textwidth}{!}{%
\begin{tabular}{@{}c|c|cc|ccc|cc|ccc|c@{}}
\toprule
\multirow{3.5}{*}{Heterogeneity} &
  \multirow{3.5}{*}{Pseudo Label} &
  \multicolumn{5}{c|}{Global memorization (server-side)} &
  \multicolumn{5}{c|}{Local memorization (client-side)} & \multirow{3.5}{*}{Test Acc.} \\ \cmidrule(l){3-12} 
 &
   &
  \multicolumn{2}{c|}{Clean label} &
  \multicolumn{3}{c|}{Noisy label} &
  \multicolumn{2}{c|}{Clean label} &
  \multicolumn{3}{c|}{Noisy label} & \\ \cmidrule(l){3-12} 
                         &                       & Correct & Wrong & Correct & Wrong & Memorized & Correct & Wrong & Correct & Wrong & Memorized \\ \midrule
\multirow{4}{*}{non-i.i.d.} & Vanilla                    & 69.10   & 30.90 & 32.96   & 34.61 & 32.43      & 95.38   & 4.62  & 12.40   & 7.10  & 80.50   &  66.54 \\
                         & Local Only  & 80.11   & 19.89 & 62.48   & 26.52 & 11.00     & 95.61   & 4.39  & 40.19   & 10.04 & 49.77  & 81.12   \\
                         & Global Only & 90.67   & 9.32  & 74.63   & 11.31 & 14.05     & 94.79   & 5.21  & 72.56   & 6.72  & 20.72  &  82.09  \\
\rowcolor{gray!20}\cellcolor{white}       & Mixture (FLR)         & 88.04   & 11.96 & 70.36   & 21.51 & 8.13      & 97.64   & 2.36  & 71.58   & 13.03 & 15.39   &  \textbf{83.04}  \\ \midrule
\multirow{4}{*}{i.i.d.}     & Vanilla                    & 90.92   & 9.08  & 16.57   & 11.16 & 72.26     & 99.96   & 0.04  & 0.00    & 0.00  & 100.00 & 67.87 \\
                         & Local Only  & 94.88   & 5.12  & 80.34   & 5.46  & 14.19      & 99.55   & 0.45  & 21.22   & 0.88  & 77.90 & 83.88  \\
                         & Global Only & 96.38   & 3.62  & 77.93   & 7.56  & 14.51     & 99.27   & 0.73  & 58.41   & 2.41  & 39.17   & 85.82  \\
\rowcolor{gray!20}\cellcolor{white}       & Mixture (FLR)         & 95.70   & 4.30  & 79.30   & 9.35 & 11.35     & 98.24   & 1.76  & 81.96   & 2.07  & 15.97   & \textbf{87.07}  \\ \bottomrule
\end{tabular}%
}
\vspace{-5pt}
\label{tab:memoriztion_stuff}
\end{table*}

\paragraph{Experimental Results} Our empirical observations indicate that FLR effectively reduces both local and global memorization when compared to models trained using the general FedAvg method, which relies solely on cross-entropy loss. As demonstrated in \autoref{fig:flr_general_memo} (a), the learning curve is smoother for FLR, and the amount of memorization is considerably lower. In addition, the model trained with FLR accurately identifies the ground truth labels of noisy samples as the rounds progress, while maintaining the memorized ratio of incorrect labels below 10\% (\autoref{fig:flr_general_memo} (b)). We employ warm-up training for regularization with server prediction during the first 50 epochs and switch to local running average prediction regularization thereafter. This transition causes the inflection point in the graph to appear shortly after warm-up completion in both \autoref{fig:flr_general_memo} (a) and (b). Although the accuracy for clean data experiences a slight decrease, the memorization of noisy data is significantly reduced.

\paragraph{Analysis according to Data Heterogeneity} \autoref{tab:memoriztion_stuff} presents the server-side and client-side memorization for both i.i.d. and non-i.i.d. settings when using local and global moving averages. In i.i.d. settings, employing local moving averages alone effectively mitigates global memorization but not local memorization. In contrast, in non-i.i.d. scenarios, using the global moving average has a more significant impact on reducing memorization. This can be attributed to the fact that when clients have similar data distributions, local information is sufficient to prevent memorization in the global model. However, in non-i.i.d. situations, relying solely on local representation is inadequate for preventing global model memorization, making the use of the global moving average more advantageous. In both cases, utilizing a mixture of local and global moving average predictions as synthetic pseudo labels produced the best results, as shown in the table's Mixture rows. Although the degree of data heterogeneity may affect the outcome, a balanced mixture of global and local predictions seems to be more effective than adopting biased pseudo labels from either side exclusively.

\begin{figure}[t]
    \centerline{\includegraphics[width=1.0\linewidth]{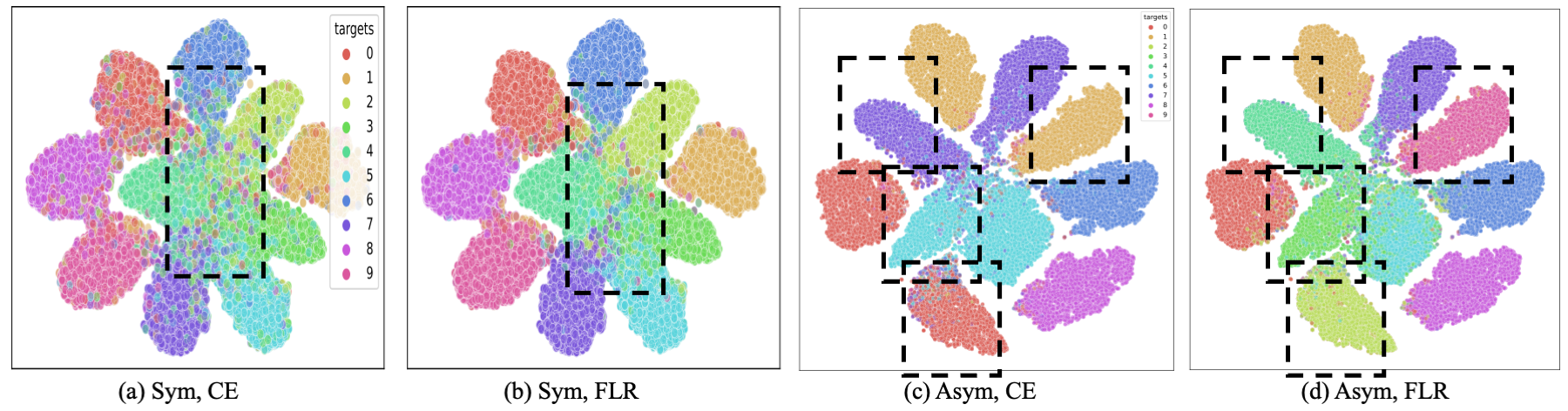}}
    \caption{t-SNE mapping view of our latent features. Colors represent the noisy label $y$ and the ground truth label $y^*$.} \label{fig:tsne_fig}
\vspace{-10pt}
\end{figure}

\paragraph{Qualitative Analysis} We can implicitly find out the quality of FLR by looking at the  color change of dots from \autoref{fig:tsne_fig}\,(e.g., (a) $\rightarrow$ (b), (c) $\rightarrow$ (d)). \autoref{fig:tsne_fig} (a) and (c) are the implicit distributions of classified noisy data through the model trained with CE, and \autoref{fig:tsne_fig} (b) and (d) represent the performance of FLR classifying them. Black dashed boxes show visible differences between them.

\subsection{Implementation}\label{subsec:implement}

\begin{figure}
    \includegraphics[width=1.0\linewidth]{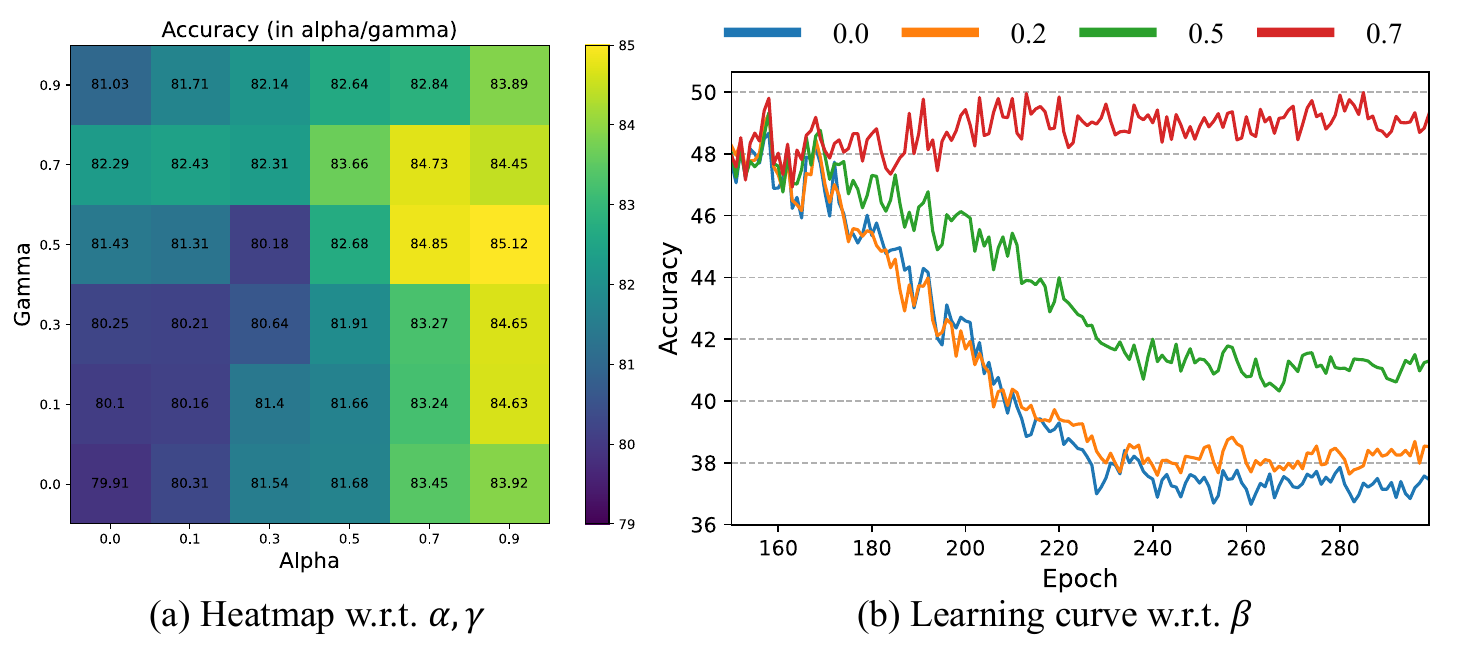}
    \caption{(a) Heatmaps with respect to $\alpha,\gamma$ at $(\rho, \tau) = (0.6, 0.5)$ on CIFAR-10, (b) learning curves according to the changes of $\beta$ at $(\rho, \tau) = (1.0, 0.5)$ on CIFAR-10. \label{fig:ablation}}
\end{figure}

\autoref{fig:overview_method} provides an illustration of how our method affects model updates. It is widely recognized that using a running average (i.e., temporal ensembling\,\cite{temporal}) is more effective than relying on static model output. As a result, the baseline version of our method follows Eq.(~\ref{eq:temporal}). We empirically investigate the effects of the components $\alpha, \beta, \gamma$ (\autoref{fig:ablation}). Insights for FLR's effective label noise handling in FL settings include:

\begin{itemize}
\item Higher values of $\alpha$ produce better results, and a certain value of $\gamma$ greater than 0 is sufficient (\autoref{fig:ablation} (a)).
\item Global memorization occurs in the latter half of learning in settings with extremely severe noise. By setting $\beta>0$, we confirm that global memorization is prevented (\autoref{fig:ablation} (b)).
\end{itemize}

\paragraph{Implications of $\alpha, \beta, \gamma$} We briefly discuss the implications of our methods:
\begin{itemize}
    \item \textbf{$\alpha$}: Balances between local and global memorization. A lower $\alpha$ leads the model to focus more on local memorization, capturing specific patterns unique to each client, but risks overfitting. A higher $\alpha$ shifts the emphasis towards global memorization, encouraging learning of universal patterns across the dataset. 
    
    \item \textbf{$\beta$}: Addresses global memorization in the server model, especially after several epochs, by employing temporal ensembling. This helps the model to maintain an overarching perspective on the learned patterns.

    \item \textbf{$\gamma$}: Aims to control local memorization. While the server model counters local memorization initially, $\gamma$ becomes essential when local memorization re-emerges at a certain convergence point. However, applying $\gamma$ too early can interfere with the learning trajectory, thus it is strategically implemented after reaching a certain level of learning.  
\end{itemize}

Expanding on the roles of $\alpha, \beta, \gamma$, we provide further insights into their effective use in FLR. The linear scheduler for $\alpha$ delicately manages the balance between local and global learning. Initially favoring local learning due to the global model's lower accuracy, $\alpha$ gradually shifts focus to global learning as noise management improves. Both $\beta$ and $\gamma$ start at zero to accommodate initial inaccuracies and increase progressively, harmonizing the blend of local and global predictions over the course of training. Our typical settings for these parameters are detailed, with further explanations provided in the Appendix.

%% file: section/3_exp.tex
\section{Experiments}\label{sec4}

\subsection{Experimental Setup}

We evaluate our methods on two standard benchmark datasets, CIFAR-10 and CIFAR-100\,\cite{cifar}, and two real-world datasets, CIFAR10N\,\cite{wei2021learning} and Clothing1M\,\cite{clothing}, with varying numbers of clients. We conduct experiments with 100 clients for CIFAR-10 and CIFAR10N, 50 clients for CIFAR-100, and 500 clients for Clothing1M. We consider both i.i.d. (CIFAR-10/100) and non-i.i.d. settings (CIFAR-10/100, CIFAR10N, Clothing1M); non-iidness is parameterized by $p$ and $\alpha_{Dir}$. As CIFAR-10/100 do not inherently contain label noise, we introduce synthetic label noise, where the level of noise is parameterized by $\rho$ and $\tau$. CIFAR10N and Clothing1M naturally contain label noise, with data samples randomly distributed among clients. $p$, $\alpha_{Dir}$, $\rho$, and $\tau$ are elaborated upon in Section~\ref{problem setup}. To contrast between CL and FL, we also compare our methods with conventional LNL methods in centralized settings, implemented following \citet{fedcorr}. In the centralized scenario, we employ a dataset that has been corrupted using the exact same scheme as in the federated scenario. Further implementation specifics are available in the Appendix.

\subsection{Results on Real World Noisy Datasets}

\begin{figure}[t]
    \centerline{\includegraphics[width=1.0\linewidth]{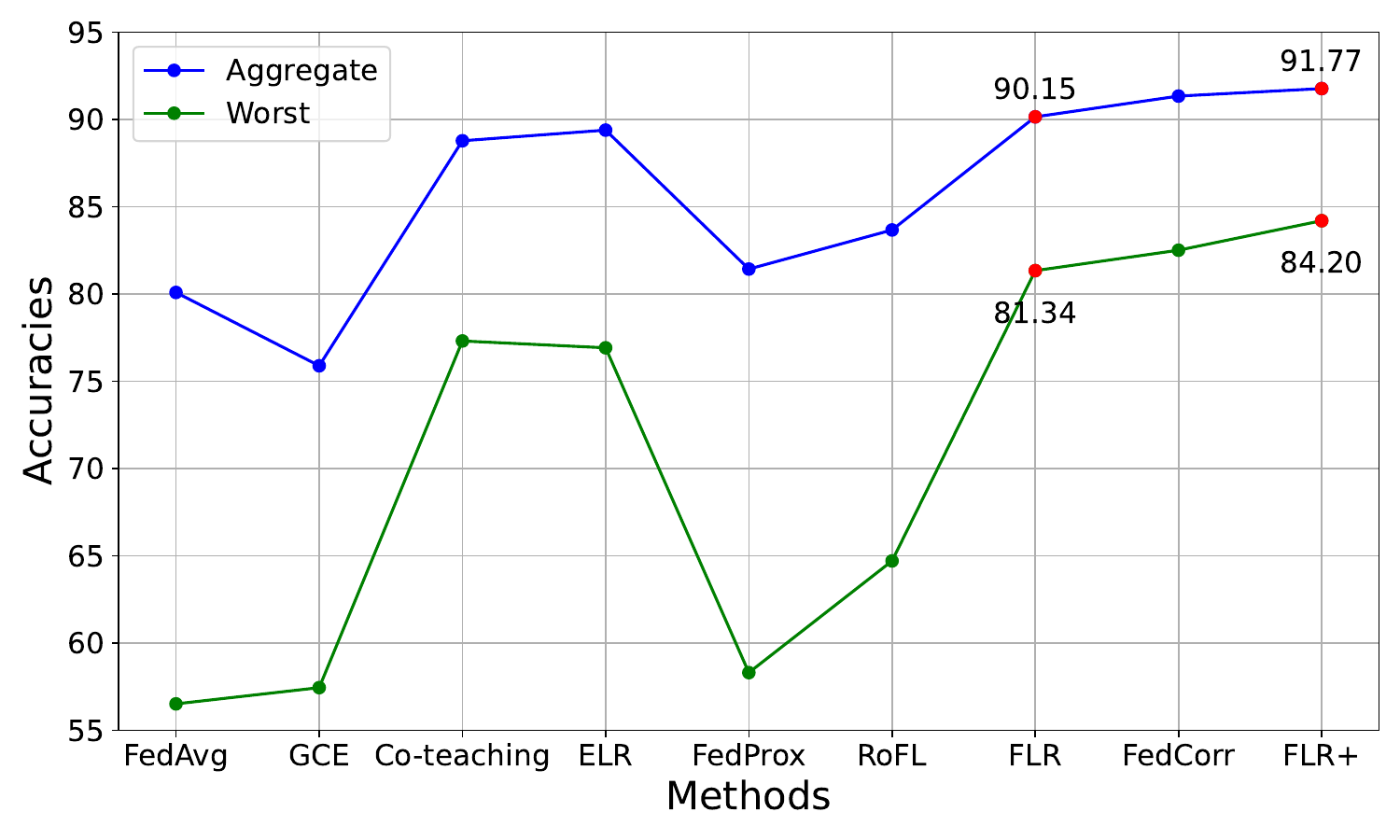}}
    \vspace{-5pt}
    \caption{Test accuracies for each method, evaluated on CIFAR10N dataset \cite{wei2021learning} in federated scenarios. ELR is implemented with FedAvg. `Aggregate' and `Worst' denote different noise type of CIFAR 10N dataset.} \label{fig:cifar10n}
\vspace{-10pt}
\end{figure}

\input{tables/clothing1m}

\begin{figure*}[t]
    \includegraphics[width=1.0\linewidth]{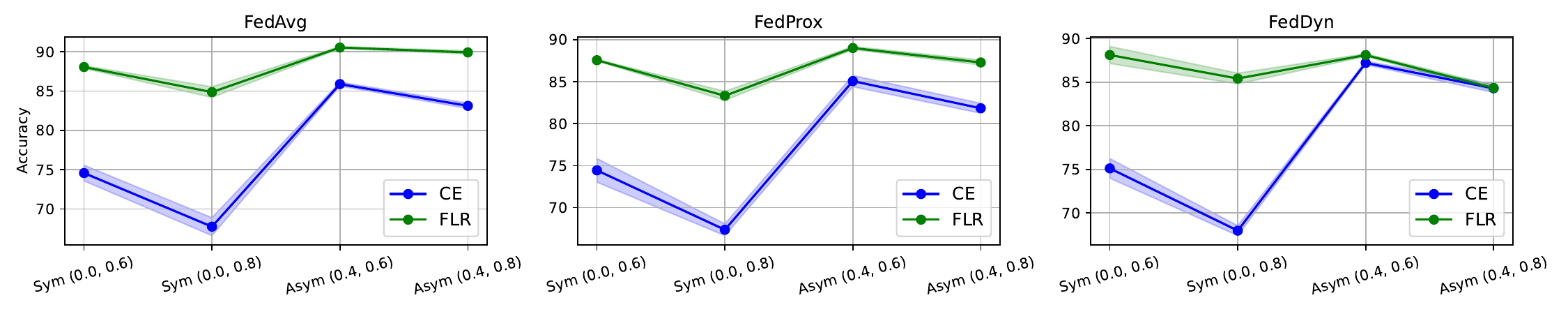}
    \caption{Collaborations with server-aware FL methods (FedProx \cite{fedprox} and FedDyn \cite{feddyn}) on CIFAR-10 with i.i.d. setting.} \label{fig:server_aware}
\end{figure*}

\paragraph{CIFAR10N} Our experiments demonstrate that FLR surpasses traditional LNL methods utilized in FL. Notably, FLR's effectiveness is further amplified when combined with FedCorr, indicating its robustness and adaptability across diverse FL settings. \autoref{fig:cifar10n} showcases FLR's significant advantage over other label regularization techniques.


\paragraph{Clothing1M} FLR consistently outperforms other LNL methods. Although the accuracy differences among the methods are modest, FLR's performance stands out. This consistency in performance is particularly noteworthy considering the use of a pre-trained model for the initial global model in these experiments. The results, detailed in \autoref{tab:clothing1M}, underscore FLR's capability to improve model accuracy in FL settings, even under conditions with pretrained models.

\subsection{Results on CIFAR Datasets}

\input{tables/cifar100}

\subsubsection{I.I.D. Setting}

Our comprehensive evaluation on CIFAR-100 under an i.i.d. setting (\autoref{tab:cifar100}) showcases FLR's remarkable performance superiority across diverse label noise settings, including both symmetric and asymmetric noise scenarios. We observe that even well-established CL methods like DivideMix \cite{dividemix}, while robust in CL, do not seamlessly translate their effectiveness into the federated context. This finding resonates with prior literature on federated learning (FedCorr \cite{fedcorr}, FedRN \cite{fedrn}). Notably, Generalized Cross Entropy (GCE) \cite{gce} exhibits strong performance, particularly in high symmetric noise scenarios, yet FLR consistently outperforms it in most cases. Combining FLR with FedCorr (i.e., FLR$^+$) yields a synergistic effect, leading to the highest test accuracies across almost all noise settings, except for the noise-free case. This combination demonstrates FLR's versatility and potential for enhancing FL frameworks under diverse noise conditions.

\begin{figure*}[t]
    \vspace{-10pt}
    \centerline{\includegraphics[width=0.85\linewidth]{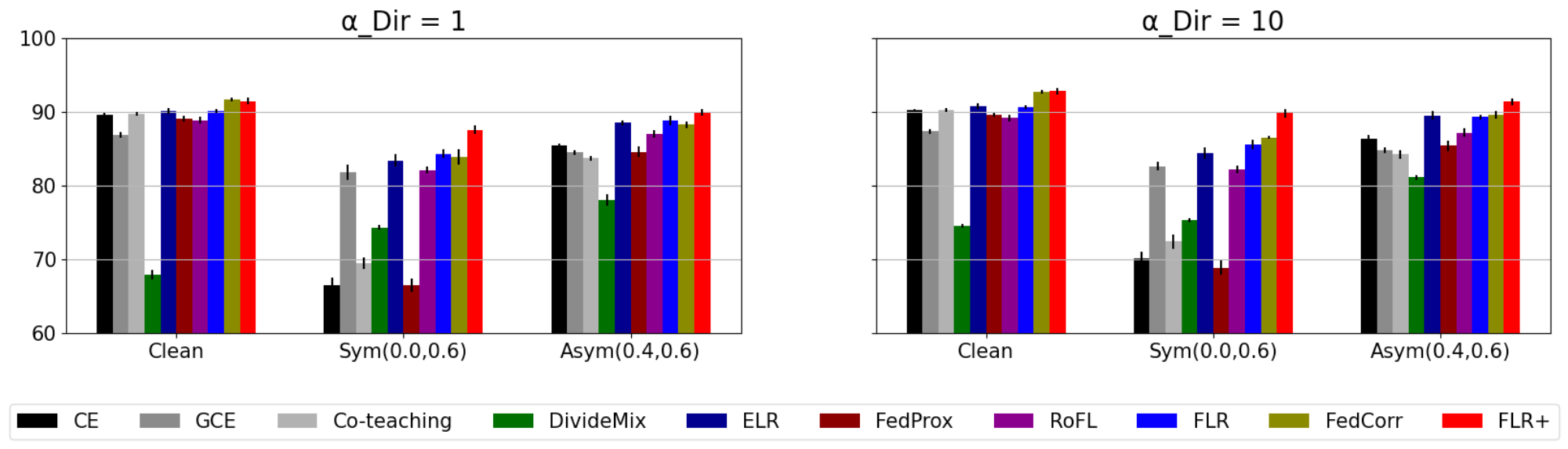}}
    \vspace{-10pt}
    \caption{Average of the best test accuracy of 5 trials for each method, on CIFAR-10 with non-i.i.d. setting by varying the concentration parameter $\alpha_{Dir}$ under different noise levels. The x-axis indicates the noise level by changing the $(\tau, \rho)$. } \label{fig:dir_fig}
\end{figure*}

\paragraph{Collaborations with Server-Aware FL Methods.}
We also delve into how FLR collaborates with server-aware FL methods. Our findings, presented in \autoref{fig:server_aware}, demonstrate that integrating FLR with these methods significantly boosts accuracy in various noisy client scenarios. This enhancement is observed consistently across different ratios of noisy clients, highlighting FLR's capability to complement server-aware strategies. This synergy is particularly crucial in federated environments where client data is diverse and often unreliable, showcasing FLR's versatility and effectiveness in improving outcomes in complex FL settings.

\subsubsection{Non-I.I.D. Setting}

\paragraph{Concentration Parameter in Dirichlet Distribution $\alpha_{Dir}$}
\autoref{fig:dir_fig} shows the results by changing the data heterogeneity $\alpha_{Dir}$ at $p=1.0$. Overall, it can be seen that the performance is lower when $\alpha_{Dir}=1$ (local data heterogeneity $\uparrow$) than when $\alpha_{Dir}=10$ (local data heterogeneity $\downarrow$). It is confirmed that FLR consistently shows better performance at most noise levels and there seems more a clear difference, especially when data heterogeneity becomes more severe.

%% file: tables/clothing1m.tex
\begin{table*}[t]
\caption[Clothing1M with IID setting]{Average and standard deviation of the best test accuracy of 5 trials for each method, evaluated on Clothing1M dataset \cite{clothing} in federated scenarios with uniformly random distributed settings. FLR$^+$ denotes the accuracy of the FLR with FedCorr framework. }
\vspace{-10pt}
\begin{center}
\resizebox{\textwidth}{!}{\begin{tabular}{ccccccccc}
	\toprule
	FedAvg  & GCE  & Co-teaching & ELR  & FedProx & RoFL & \cellcolor{gray!20}FLR & FedCorr & \cellcolor{gray!20}FLR$^+$\\
	\midrule
	67.73$\pm$0.18 & 63.57$\pm$2.85 &  67.67$\pm$0.10 & 68.68$\pm$0.16  & 66.99$\pm$0.13 & 67.68$\pm$0.17 & \cellcolor{gray!20}{69.76$\pm$0.19} & {69.93$\pm$0.11} &   \cellcolor{gray!20}\textbf{70.11$\pm$0.14}\\
	\bottomrule
\end{tabular}}
\end{center}
\vspace{-10pt}
\label{tab:clothing1M}
\end{table*}

%% file: tables/cifar100.tex
\begin{table}[t]
\addtolength{\tabcolsep}{-4pt}\tiny
\caption{Average and standard deviation of the best test accuracy of 5 trials for each method, on CIFAR-100 with i.i.d. setting and different noise levels.  
}
\centering
\begin{tabular}{ll|cc|cc|c}
	\toprule
	\multirow{4}{*}{Setting} & \multirow{4}{*}{Method} & \multicolumn{4}{c|}{Symmetric} & \multicolumn{1}{c}{Asymmetric} \\
	\cmidrule{3-7}
	&  & \multicolumn{2}{c|}{$\tau=0.0$} & \multicolumn{2}{c|}{$\tau=0.5$} & \multicolumn{1}{c}{$\tau=0.4$} \\
	\cmidrule{3-7}
	&  & $\rho=0.6$ & $\rho=0.8$ & $\rho=0.6$ & $\rho=0.8$ & $\rho=0.6$  \\
	\midrule
	Centralized
		& CE  & 50.62$\pm$1.42 & 46.30$\pm$1.88 & 43.02$\pm$1.08 & 32.92$\pm$1.26 & 55.74$\pm$0.12 \\
		& GCE  & 61.69$\pm$0.77 & 59.06$\pm$0.92 & 57.25$\pm$0.58 & 48.91$\pm$1.32 & 62.48$\pm$0.51  \\
		& Co-teaching  & 42.68$\pm$1.24 & 38.45$\pm$1.88 & 36.38$\pm$0.55 & 27.18$\pm$0.81 & 41.90$\pm$1.33  \\
		& DivideMix$^\dagger$  & \textbf{69.05$\pm$0.90} & \textbf{65.86$\pm$1.34} & \textbf{64.64$\pm$0.65} & \textbf{54.12$\pm$0.71} & \textbf{69.75$\pm$0.40}  \\
 		& ELR  & 61.26$\pm$1.59 & 56.58$\pm$2.09 & 53.09$\pm$1.21 & 39.71$\pm$1.10 & 69.57$\pm$0.16  \\
	\specialrule{.1em}{.05em}{.05em} 
	Federated
		& FedAvg  & 42.99$\pm$1.29 & 36.68$\pm$2.02 & 34.53$\pm$1.12 & 25.48$\pm$0.80 & 51.79$\pm$0.32  \\		
		& GCE  & 48.51$\pm$1.37 & 43.18$\pm$1.54 & 40.55$\pm$1.65 & 30.96$\pm$1.14 & 54.60$\pm$0.48  \\
		& Co-teaching  & 44.74$\pm$1.23 & 38.81$\pm$1.75 & 36.55$\pm$0.83 & 27.61$\pm$0.97 & 53.30$\pm$0.82 \\
		& DivideMix$^\dagger$  & 54.48$\pm$0.98 & 52.26$\pm$0.65 & 49.37$\pm$1.17 & {43.64$\pm$0.92} & 55.46$\pm$0.15  \\
		& ELR  & 51.02$\pm$1.58 & 43.47$\pm$2.00 & 40.58$\pm$1.71 & 28.50$\pm$1.03 & 56.98$\pm$0.67  \\
  		& FedProx  & 40.61$\pm$1.47 & 34.36$\pm$1.95 & 32.71$\pm$0.99 & 24.04$\pm$0.85 & 49.11$\pm$0.68  \\
    	& RoFL  & 43.44$\pm$1.36 & 36.61$\pm$2.12 & 38.61$\pm$1.13 & 26.53$\pm$1.07 & 51.83$\pm$0.65  \\
		\rowcolor{gray!20}\cellcolor{white}  & FLR  & {59.13$\pm$1.61} & {54.18$\pm$1.87} & {51.12$\pm$1.46} & 40.15$\pm$1.34 & {65.21$\pm$0.56}  \\
	
		& FedCorr  & 53.23$\pm$1.81 & 46.25$\pm$1.95 & 44.13$\pm$0.61 & 31.43$\pm$1.54 & 59.61$\pm$0.56  \\
            \rowcolor{gray!20}\cellcolor{white}  & FLR$^+$  & \textbf{67.85$\pm$0.91} & \textbf{64.12$\pm$2.01} & \textbf{64.64$\pm$0.38} & \textbf{54.10$\pm$1.01} & \textbf{69.52$\pm$0.71}  \\
	\midrule
	\bottomrule
	\multicolumn{4}{l}{$\dagger$\,: results with 3 trials}
\end{tabular}
\vspace{-10pt}
\label{tab:cifar100}
\end{table}

%% file: section/4_related_work.tex
\section{Related Work}

\subsection{Learning with Noisy Label}

Label noise in supervised learning can significantly reduce the generalization capability of deep neural networks (DNNs), leading to the development of various methods for robust training\,\cite{nll}. Recent progress in this field has included designing robust architectures\,\cite{can,rcnn,ra}, loss functions, and regularization techniques\,\cite{ls,mixup,elr,nls}, as well as methods for loss adjustment\,\cite{boot,d2l,adacorr} and sample selection\,\cite{cot,dividemix,kim2021fine}. However, most of them have been proven to work only under CL settings. 

One of the major challenges in dealing with label noise is the DNN's ability to memorize unreliable data, including random data or noisy labels, a.k.a., memorization phenomenon. This memorization phenomenon is observed to occur during the training phase, where DNNs first learn correctly labeled data and then gradually memorize the wrong labels\,\cite{memo1-1, memo1-2, memo2}. To prevent the memorization phenomenon, some approaches have focused on stopping training before memorization becomes severe\,\cite{es,gdes,ies}, and others have utilized regularization in the early-learning phase\,\cite{elr,rel}. However, their effectiveness in the FL setting has yet to be fully explored.

\vspace{-5pt}
\subsection{Federated Learning with Noisy Label}

Federated Learning with Noisy Labels (FNL) is a scenario where the clients have label noise in their local datasets. This problem is particularly prevalent in FL, where data is decentralized and collected from multiple clients with varying levels of label noise. Training a model with noisy labels can be detrimental as over-parameterized neural networks are prone to fitting the training dataset, leading to a memorization problem. Existing FNL methods address this issue by using techniques such as sample selection, robust aggregation, label correction, and robust loss functions. For instance, FedCorr\,\cite{fedcorr} corrects label noise in multiple stages, such as preprocessing, fine-tuning, and regular training, and FedRN\,\cite{fedrn} robustly aggregates local models by weighting them based on the estimated label noise level. However, some studies have limitations, such as relying on supervision\,\cite{focus, tuor} or not analyzing results in non-i.i.d. settings and situations with heterogeneous noise levels\,\cite{lsr, fedrn}.

%% file: section/5_conclusion.tex
\section{Conclusion}

This paper presents Federated Label-Mixture Regularization (FLR), an innovative method designed to tackle label noise in FL environments. FLR is built upon our observation that FL models are susceptible to both local and global memorization issues. By creating novel pseudo labels, which are a combination of global running average predictions and local running average predictions, FLR effectively mitigates these memorization issues. Through comprehensive experiments, FLR demonstrates a marked improvement in global model accuracy across both i.i.d. and non-i.i.d. data settings. While it shows promising results, we acknowledge that FLR may not be optimally suited for all types of label noise. Future research could delve into exploring complementary regularization techniques to address various noise structures. The potential impact of FLR on real-world applications, especially in sectors like healthcare and financial services where data privacy is paramount, is significant.

%% file: appendix/app1.tex
\appendix
\onecolumn

\section{Overview of Appendix}
We present additional details, results, and experiments that are not included in the main paper.

\section{Ethics Statement}
As we advance the field of Federated Learning (FL) with our novel Federated Label-mixture Regularization (FLR) method, it is crucial to consider its ethical implications. We address potential concerns in the areas of privacy, fairness, environmental impact, and potential misuse.

\paragraph{Privacy and Data Security} FLR, like other FL methods, is designed to maintain the privacy of clients' data by performing computations locally and only sharing model updates. This approach inherently mitigates privacy concerns arising from the transmission of raw data. However, potential risks persist with the possibility of model inversion attacks or other adversarial actions. Future work must continue to focus on strengthening the robustness of FL methods against such threats.

\paragraph{Fairness} FLR could potentially exacerbate or mitigate existing fairness issues in FL depending on how it is deployed. On one hand, if FLR is used predominantly with data from certain demographics, it may introduce bias into model predictions. On the other hand, FLR's ability to handle noisy labels may allow for the inclusion of previously excluded data, promoting a more diverse and fair model.

\paragraph{Environmental Impact} FLR, like other FL methods, reduces the need for centralized data storage and computation, potentially lowering the carbon footprint associated with these processes. However, the energy consumption of local computation and communication for model updates must be carefully managed to ensure environmental sustainability.

\paragraph{Potential Misuses} While FLR is designed to improve the robustness of FL methods against noisy labels, it could potentially be misused to intentionally introduce bias or misinformation into models. For instance, malicious actors could leverage our method to facilitate their objectives by manipulating the noise in labels. Hence, it is crucial to develop safeguards against such misuse.

In conclusion, while FLR contributes positively to the FL field by offering a solution to handle noisy labels, it is essential to apply it thoughtfully, considering its potential ethical implications. We encourage ongoing dialogue and scrutiny to ensure that its deployment aligns with principles of privacy, fairness, and social responsibility.

\section{Limitations}
While our FLR method presents several noteworthy advancements in handling noisy labels in FL, it is crucial to acknowledge the limitations that accompany it:

\begin{enumerate}
    \item \textbf{Performance with Highly Imbalanced Data:} Although our FLR method demonstrates robustness in diverse data heterogeneity scenarios, its efficacy in situations with severe data imbalance across clients is an area that requires further investigation. Data imbalance here refers to the uneven distribution of data classes across different clients. For example, one client might have a majority of samples from a particular class while another client might have a very few samples of that class. This severe skewness in data distribution can lead to biased learning, where the model performs well on classes with more samples but poorly on classes with fewer samples. Although FLR has mechanisms to mitigate the effects of data heterogeneity, it is not explicitly designed to handle extreme data imbalance.

    \item \textbf{Computational Overhead:} While the FLR method is effective in improving the model's robustness against noisy labels, it also introduces additional computational overhead. This is primarily due to the calculation of local and global moving averages, which are central to the FLR method. The moving average calculations require additional computation and memory resources, which might not be suitable for resource-constrained devices often participating in FL. This could potentially limit the scalability and applicability of our method in real-world FL scenarios, particularly where computational resources are a constraint. Therefore, optimizing the computational efficiency of FLR without compromising its effectiveness against noisy labels is an important direction for future research.

    \item \textbf{Noise Type Sensitivity:} FLR has been specifically developed to tackle the challenge of label noise in FL. This label noise can occur due to various reasons, such as human errors during data labeling or miscommunication during data transfer. However, in real-world scenarios, there are other types of noise that can equally affect the learning process. For instance, feature noise refers to inaccuracies or inconsistencies in the input data, which might be caused by faulty sensors or measurement errors. Model noise, on the other hand, is related to the discrepancies between the true data generating process and the model assumptions. Since our FLR method primarily focuses on label noise, it might not exhibit the same level of efficiency when dealing with feature or model noise. Future iterations of FLR could explore these areas to provide a more comprehensive solution to noise handling in FL.
\end{enumerate}

\section{Future Directions}
Despite the aforementioned limitations, the FLR method opens up several promising avenues for future research:

\begin{enumerate}
    \item \textbf{Handling Other Noise Types.} Future work can look into extending the FLR method to handle other types of noise efficiently. This would make it a more comprehensive solution for real-world federated scenarios where different types of noise coexist.

    \item \textbf{Adaptation for Imbalanced Data.} Investigating and enhancing the performance of FLR with highly imbalanced data distribution would be a valuable future direction. Techniques like adaptive resampling or cost-sensitive learning could be integrated with our method to tackle this challenge.

    \item \textbf{Applications on Natural Language Processing:} Despite the emergent ability of Large Language Models (LLM), they also struggle with the privacy issues. Employing LLM training in federated scenario may meet noisy labels and texts with high probability, and thus our work can be a good solution in this area.

    \item \textbf{Optimization of Computational Efficiency.} Future research could also focus on optimizing the computational efficiency of the FLR method. Reducing the computational overhead without compromising the robustness against noisy labels would make our method more practical for real-world FL scenarios.

    \item \textbf{Robustness Against Adversarial Attacks.} As the FL domain evolves, adversarial attacks pose an increasing threat to model robustness. Future work could explore how to bolster the FLR method (and FL methods, in general) to ensure robustness against adversarial attacks.
    
\end{enumerate}

By addressing these limitations and exploring these future directions, we can continuously refine and evolve the FLR method to better serve the ever-growing demands of FL.

%% file: appendix/app2.tex
\subsection{Previous Works for FNL}
There are several works trying to solve FNL, listed in \autoref{tab:previous_NFL}.

\input{tables/previous_nfl.tex}

\subsection{Adverarial Clients in FL}
Meanwhile, there exist some studies\,\cite{arfl,poison,aaa,sageflow} to combat adversarial attacks on federated systems. Among them, data poisoning attack might seem similar to FNL; few malicious clients aim to break the global model by sending local models trained with mislabeled data. However, FNL is distinct from it since every client can have some degree of label noises in FNL. From this perspective, we do not consider such works in our study.

\section{Experimental Settings}
The common settings are used in this paper for all baseline experiments on each datasets\,(\autoref{tab:datasets}).
\input{tables/datasets}

\subsection{Heterogeneous Data Distribution Settings on CIFAR-10}
\begin{figure}[t]
    \centerline{\includegraphics[width=0.75\linewidth]{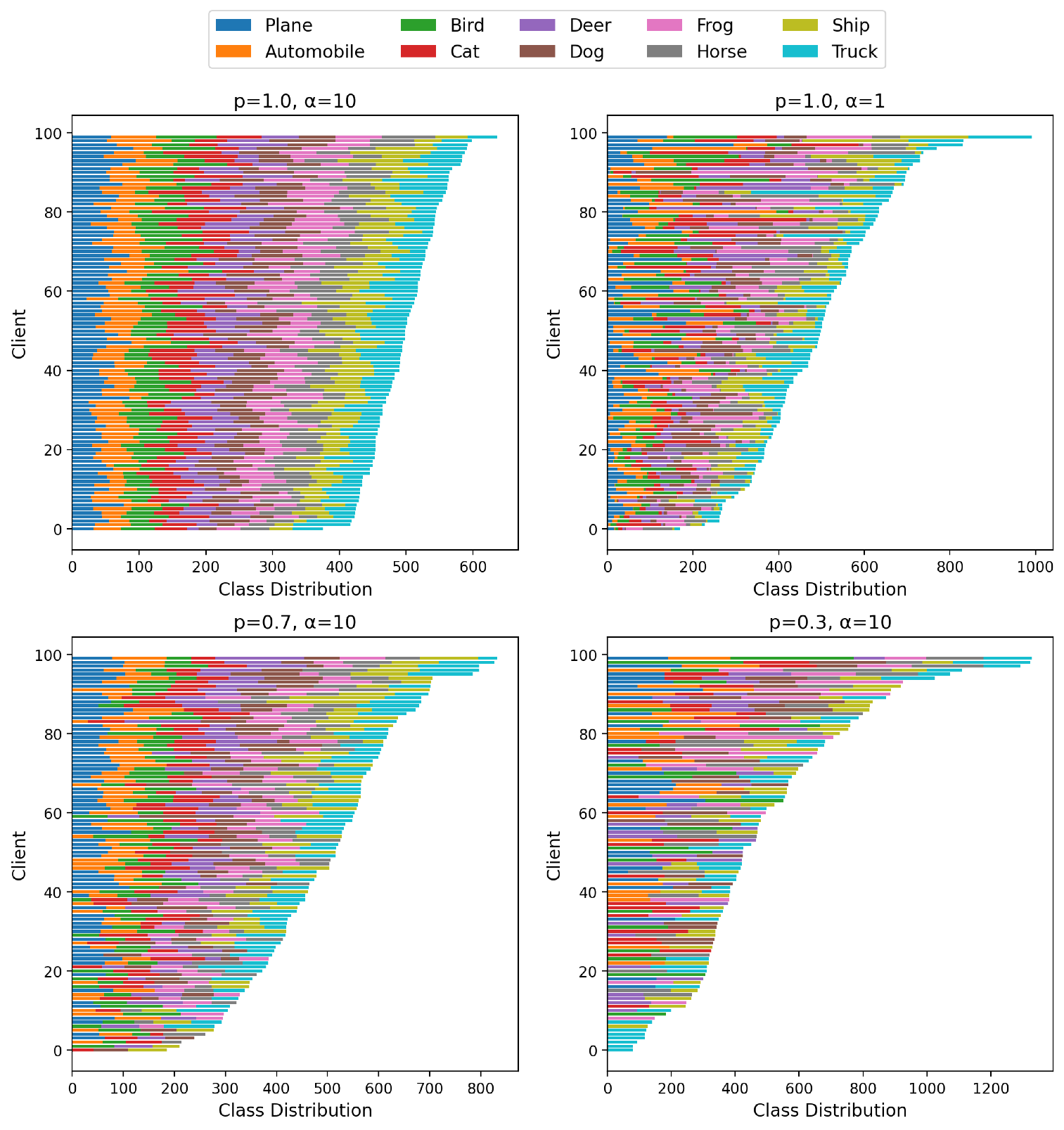}}
    \caption[CIFAR-10 with data heterogeneity]{Heterogeneous data distribution settings on CIFAR-10.
    } \label{fig:data_dist}
\end{figure}

\autoref{fig:data_dist} shows how we partition the CIFAR-10 dataset among the client's local datasets.

\subsection{Heterogeneous Data Distribution Settings on CIFAR-100}
\begin{figure}[h!]
    \centerline{\includegraphics[width=0.8\linewidth]{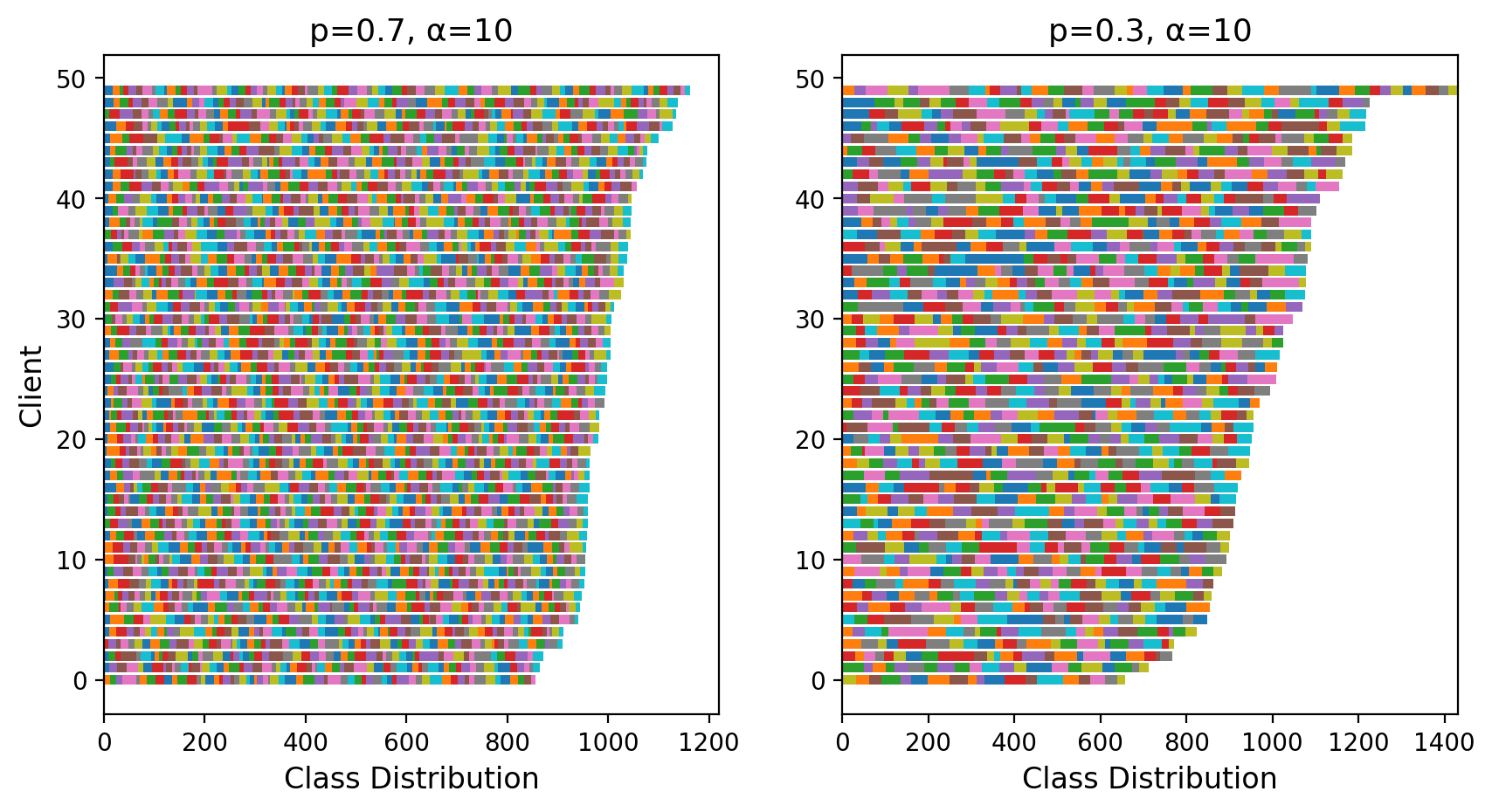}}
    \caption[CIFAR-100 with data heterogeneity]{Data distributions of non-i.i.d. settings on CIFAR-100.
    } \label{fig:data_dist_cifar100}
\end{figure}

\autoref{fig:data_dist_cifar100} shows how we partition the CIFAR-100 dataset among the client's local datasets.

\subsection{Noise Transition Matrix}
\begin{figure}[h!]
  \centering
  \begin{subfigure}[b]{0.49\linewidth}
    \includegraphics[width=\linewidth]{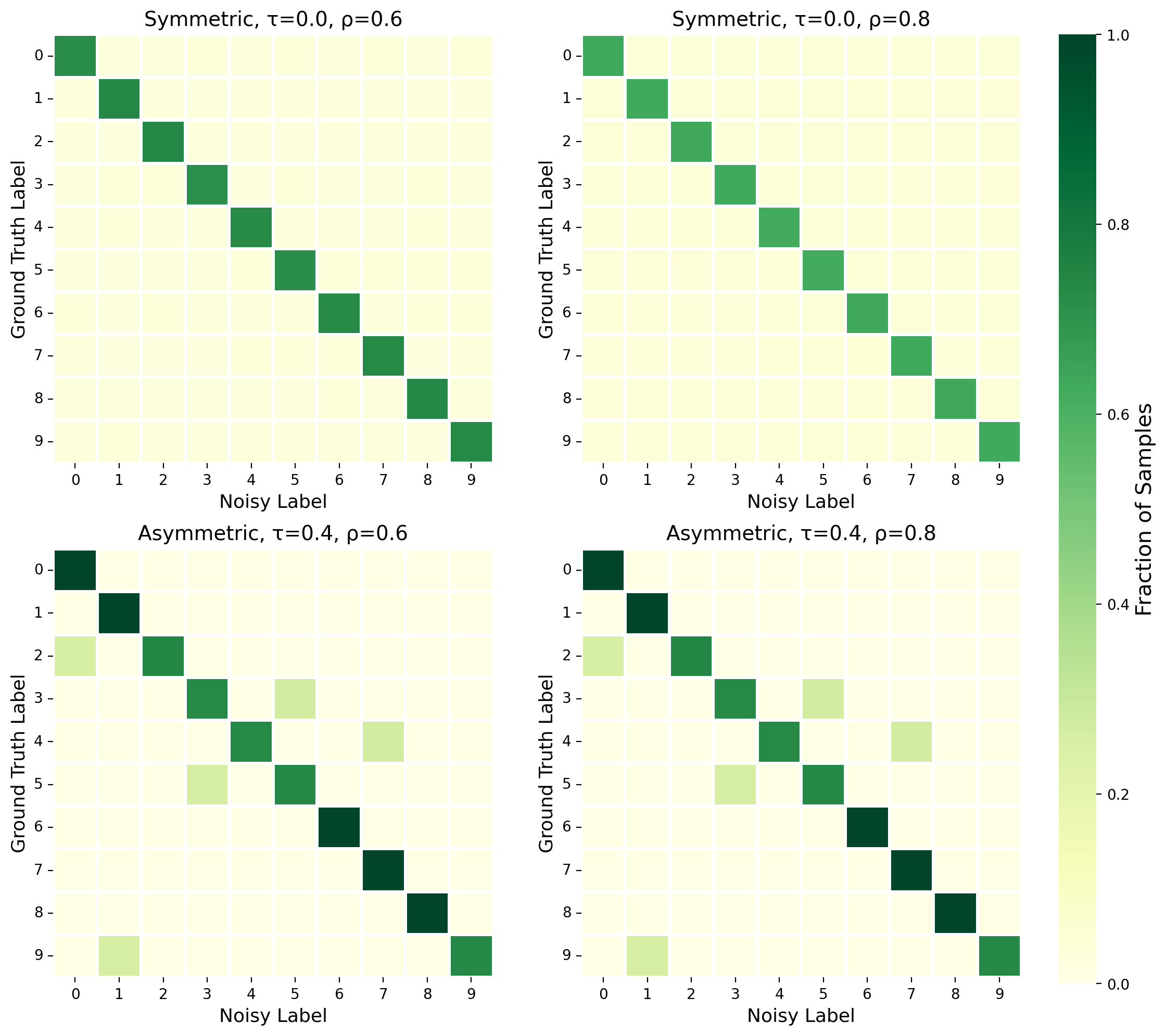}
    \caption{Noise transition matrix for CIFAR-10.}\label{fig:tm_cifar10}
  \end{subfigure}
  \begin{subfigure}[b]{0.49\linewidth}
    \includegraphics[width=\linewidth]{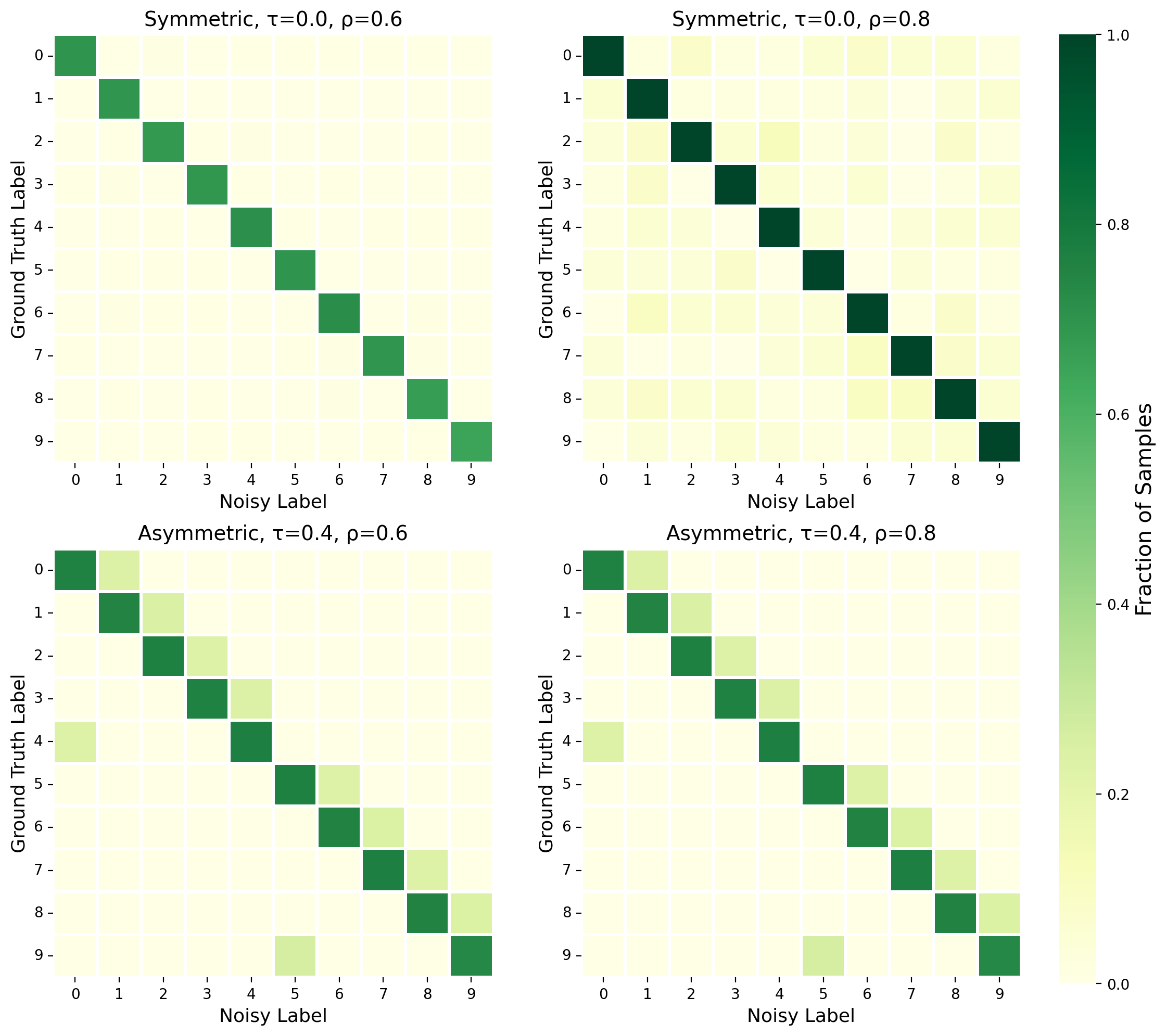}
    \caption{Noise transition matrix for CIFAR-100.} \label{fig:tm_cifar100}
  \end{subfigure}
  \caption{Noise transition matrices.}
\end{figure}

\autoref{fig:tm_cifar10} and \autoref{fig:tm_cifar100} illustrate the practical transition matrices from our implementation, which show how labels are reassigned based on their ground truth labels.

\subsection{Pseudocode for FLR Implementation}

Algorithm \ref{al:flr} provides a detailed procedure for implementing the Federated Label-mixture Regularization (FLR) method, a strategy designed to tackle noisy labels within a FL environment.

The algorithm is structured into two distinct phases. The first phase serves as a warm-up round, applying the standard Federated Averaging (FedAvg) method utilizing standard Cross-Entropy (CE) loss. This phase forms a baseline global model trained on noisy labels. The main steps are as follows:

\begin{itemize}
    \item \textbf{Initialization:} The server model parameters, denoted as $\mathcal{\theta}_{server}$, are initialized with randomly chosen parameters, $\mathcal{\theta}_{0}$.
    \item \textbf{Training Iterations:} A loop for each epoch initiates, where a subset of clients $S^e$ are selected for training. For every chosen client, they start with a model whose parameters match those of the server model ($\mathcal{\theta}_{k} = \mathcal{\theta}_{server}$). The model parameters are updated through local gradient descent steps on their individual datasets, applying the standard CE loss.
    \item \textbf{Aggregation:} Lastly, the server gathers the client models into a global model by computing a weighted average. The weight for each client model is proportional to their dataset size.
\end{itemize}

In the second phase, the FedAvg procedure is augmented with the proposed FLR loss. The steps are akin to phase one, but with the following enhancements:

\begin{itemize}
    \item \textbf{Running Average Predictions:} For each mini-batch in a client's dataset, both global and local running average predictions are computed. The global running average predictions ($\vs_{k_i}$) incorporate the server model's predictions, whereas the local running average predictions ($\vm_{k_i}$) utilize the predictions of the client's own model.
    \item \textbf{Mixed Labels:} A mixture label, $\vt_{k_i}$, is formed by combining both global and local running average predictions. This facilitates an effective correction of noisy labels.
    \item \textbf{Client Update with FLR:} The client models are updated via gradient descent, applying the proposed FLR loss. The FLR loss consists of the standard CE loss and a regularization term, which is based on the distance between the current model's prediction and the mixture label.
\end{itemize}

The delineated algorithm offers a comprehensive blueprint for implementing FL in the presence of noisy labels, effectively handling data heterogeneity and varying label noise across different clients.

\newpage
\subsection{Implementation Details} 
Our implementation details for the methods are as follows:
\begin{itemize}
    \item FedProx\,\cite{fedprox} requires one coefficient $\mu$ for its proximal term. We chose $\mu=0.001$ for our experiments.
    
    \item FedCorr\,\cite{fedcorr} has a lot of hyperparameters. Among them, the number of rounds in finetuning and usual FL stage are set to be (300, 300) for CIFAR-10, (300, 400) for CIFAR-100, and (50, 50) for Clothing1M. We followed the values of original paper for the rest of hyperparameters.

    \item GCE\,\cite{gce} requires two hyperparameters, $q$ and $\kappa$. We selected $q=0.7, \kappa=0.1$ for CIFAR-10 and Clothing1M, and $q=0.1, \kappa=0.3$ for CIFAR-100 in our experiments.
    
    \item Co-teaching\,\cite{cot} uses two models. In our implementation of Co-teaching for FL, we maintain two global models $w_1$ and $w_2$. When each round starts, both of them are given to each client as the local models $w_{1k}$ and $w_{2k}$, and each client trains them with Co-teaching process. After the local updates are done, the global models are updated by global aggregation respectively; $\{w_{1k}\}\to w_1$ and $\{w_{2k}\}\to w_2$.
    
    \item DivideMix\,\cite{dividemix} for FL is implemented in the same way as in the Co-teaching case. Regarding hyperparameters, we set 0.5 for the sharpening temperature, and 25 for the unsupervised loss coefficient.
    
    \item Early Learning Regularization (ELR)\,\cite{elr} requires two coefficients $\lambda$ and $\beta$ for its regularization term. After many trials of grid search, we found that $\lambda=4.0$ with $\beta=0.5$ for symmetric noise and $\beta=0.9$ for asymmetric noise works better than any other hyperparameter values that we tested, in most of our experimental settings. We used $\lambda=4.0$ and $\beta=0.5$ also for Clothing1M.
    
    \item FLR require several coefficients $\lambda$, $\alpha$, $\beta$, and $\gamma$, and has the option of using linear scheduling for $\alpha$. $(\alpha, \beta, \gamma)$ is fixed to (0.9, 0.7, 0.5), and other settings are specified in \autoref{tab:flr}.
    
    \item FLR$^+$ is used $\alpha_M=1$ (beta distribution parameter used for Mixup augmentation) when coupled with FedCorr framework, and $\alpha_M=4$ for DivideMix.
\end{itemize}

For learning strategy, we implement the training as follows:
\vspace{-5pt}
\begin{equation*}
    \alpha(r) = \alpha \times \frac{r}{R}, \quad
    \beta(r) = \begin{cases}
      0 & \text{if } r < \frac{R}{2}\\
      \beta & \text{else}
    \end{cases}, \quad
    \gamma(r) = \begin{cases}
      0 & \text{if } r < R_w\\
      \gamma & \text{else}
    \end{cases}
\end{equation*}

where $R_w$ denotes the epoch for warmup training, which is set to 50. For the methods developed in CL settings such as Mixup\,\cite{mixup}, GCE\,\cite{gce}, Co-teaching\,\cite{cot}, DivideMix\,\cite{dividemix}, and ELR\,\cite{elr}, we just follow the implementation of their original papers.

\input{tables/flr}

\subsection{Comparison with Other Calibration Methods}

Our adaptive regularization, especially ER, slightly overlaps with calibration methods, which have been extensively studied in CL. We compare ER with some confidence-related previous methods:
\begin{itemize}
    \item  Entropy Minimization\,(EM)\,\cite{em} requires one coefficient for its regularization term. We set the coefficient to be 4.
    
    \item Label Smoothing\,(LS)\,\cite{ls} and Negative Label Smoothing\,(NLS)\,\cite{nls} requires one smoothing parameter. We set its value to be 0.1 for LS, and -0.05 for NLS.
    
    \item Temperature Scaling\,(TS)\,\cite{calibration} requires sharpening temperature. We set the value to be 4.0.

    \item  Entropy Regularization (ER):  In scenarios where $\alpha=\gamma=0$, the regularization aligns closely with the entropy minimization term\,\cite{em}.
\end{itemize}

\autoref{tab:er} shows the results of ER comparing with Entropy Minimization\,(EM)\,\cite{em}, Label Smoothing\,(LS)\,\cite{ls}, Negative Label Smoothing\,(NLS)\,\cite{nls}, and Temperature Scaling\,(TS)\,\cite{calibration}. It is observed that even simple entropy penalization for each instance with either EM or ER can make the model to be robust towards noisy labels.

\input{tables/er}

\section{Gradients of the Adaptive Regularization}

\textbf{Lemma.} (\textcolor{black}{Gradient of the Regularization}). The gradient of the loss is defined in \eqref{gradient}. Similar to Liu et. al\,\cite{elr}, we denote the federated adaptive regularization by 
\begin{equation}
    R(\theta) = \log\left(1- \langle\vp_{k_i}\cdot \vt_{k_i}\rangle\right) 
\end{equation}

The gradient of $R$ is 
\begin{equation}
    \nabla R(\theta) = \frac{\nabla \left(1- \langle\vp_{k_i}\cdot \vt_{k_i}\rangle\right) }{1- \langle\vp_{k_i}\cdot \vt_{k_i}\rangle}
\end{equation}

It can be expressed with the probability estimate in terms of the softmax function and the deep-learning mapping $N_x(\theta)$, $\vp_{k_i} = \frac{\mathbf{\text{exp}(N_x(\theta)_c)}}{\sum_c \text{exp}(N_x(\theta)_c)}$, where the bolded numerator is a softmax probability vector. Substituting and solving this, we get

\begin{equation}
\begin{aligned}
    \nabla R(\theta) & = \frac{-\nabla N_x(\theta)}{1- \langle\vp_{k_i}\cdot \vt_{k_i}\rangle} (\vp_{k_i} \odot \vt_{k_i} - \langle \vp_{k_i} \cdot \vt_{k_i}\rangle \cdot \vp_{k_i})\\
    & =  \frac{-\nabla N_x(\theta)}{1- \langle\vp_{k_i}\cdot \vt_{k_i}\rangle} 
    \begin{bmatrix}
        \vp_{k_i}^{(1)} \cdot \sum_{r=1}^C (\vt_{k_i}^{(r)} - \vt_{k_i}^{(1)})\vp_{k_i}^{(r)} \\
        \vdots \\
        \vp_{k_i}^{(C)} \cdot \sum_{r=1}^C (\vt_{k_i}^{(r)} - \vt_{k_i}^{(C)})\vp_{k_i}^{(r)} 
    \end{bmatrix}
\end{aligned}
\end{equation}

\newpage
\section{\textcolor{black}{Learning Curves Across Different Data Heterogeneity}}

\begin{figure}[h!]
  \centering
  \begin{subfigure}[b]{0.49\linewidth}
    \includegraphics[width=\linewidth]{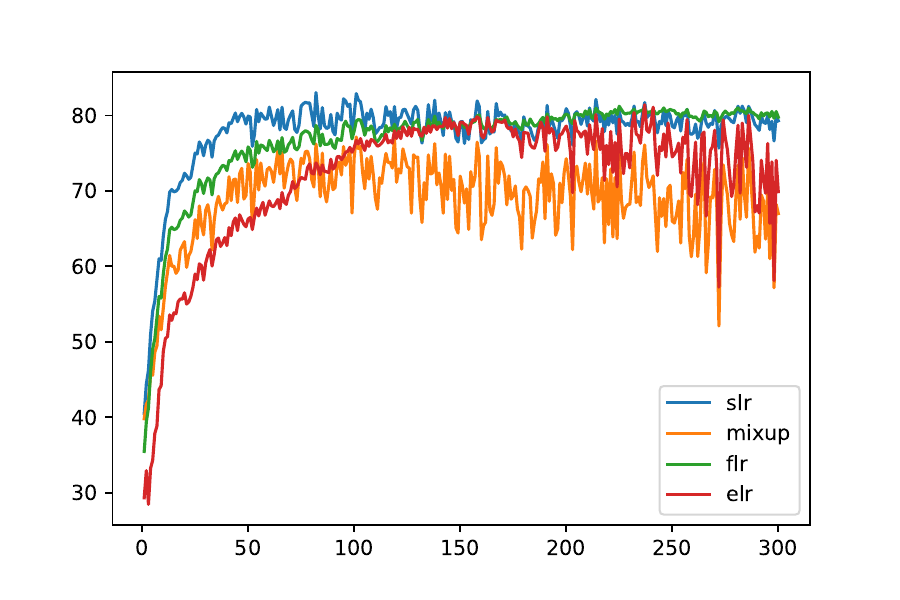}
    \caption{I.I.D.}
  \end{subfigure}
  \begin{subfigure}[b]{0.49\linewidth}
    \includegraphics[width=\linewidth]{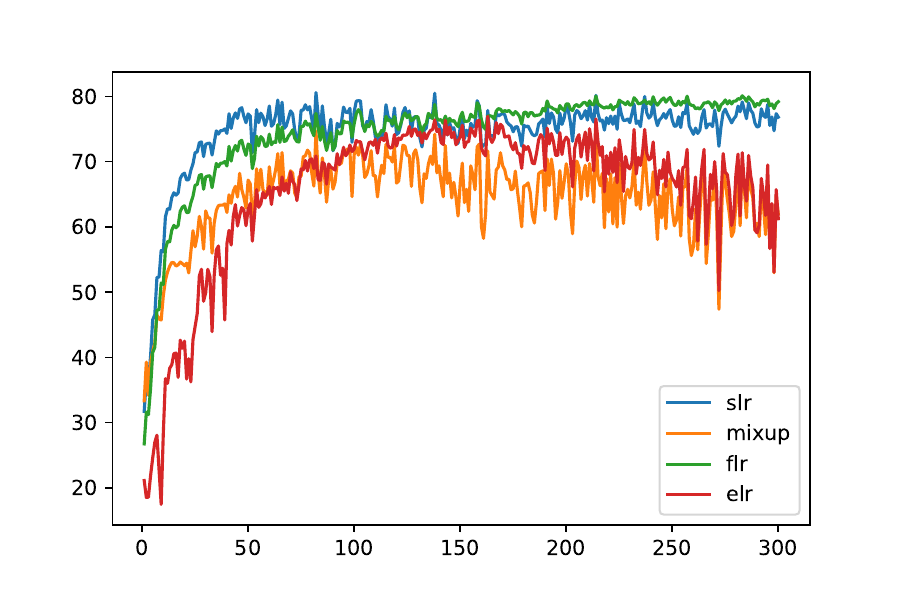}
    \caption{Non I.I.D.} 
  \end{subfigure}
  \caption{(a) CIFAR-10 of the i.i.d. setting, with symmetric noise of $(\rho,\tau)=(1.0,0.0)$, (b) CIFAR-10 of the non-i.i.d. setting, with symmetric noise of $(\rho,\tau)=(1.0,0.0)$}\label{learning_curve}
\end{figure}

Building on the equation~\ref{eq:temporal}, other regularizations are formulated according to the uses of $\alpha, \beta, \gamma$:

\begin{itemize}
    \item ELR\,\cite{elr}: This method is a specific case of FLR where $\alpha=0, \gamma>0$ in federated settings. The significant advantage of ELR is that it does not require additional storage for $\vs_{k_i}$.
    
    \item SLR: This is another special case of FLR, specifically when $\alpha=1$, where local predictions are directly penalized by global predictions. Although similar to general distillation, SLR differs in its use of the logarithmic function rather than the Kullback-Leibler function\,\cite{hinton2015distilling}. In FL, other methods like FedProx\,\cite{fedprox} and FedDyn\,\cite{feddyn} improve generalization against data heterogeneity by reducing weight divergence between the local and global model. However, our approach is distinct in that it directly mitigates divergence between predictions, rather than model parameters.
\end{itemize}

\textcolor{black}{In this section, we present the learning curves over 300 epochs for four different strategies: ELR, SLR, Mixup, and FLR. The curves are plotted under both the i.i.d. and non-i.i.d. settings.}

\textcolor{black}{Our observations indicate that both FLR and SLR generally perform well, with FLR consistently exhibiting superior performance\,(\autoref{learning_curve}). This underlines the effectiveness of our proposed FLR method in combating the effects of noisy labels in the FL environment.}

\textcolor{black}{On the other side, as the number of epochs increases, ELR and Mixup's performance tends to deteriorate. This is particularly evident in their generalization capabilities, which decrease due to memorization. This memorization effect is particularly evident when these methods are exposed to prolonged training, which is a common situation in a real-world scenario.}


%% file: tables/previous_nfl.tex
\begin{table}[h]\small
\caption[Previous works in FNL]{Comparison of previous FNL methods according to several properties: (P1) sample selection (P2) robust aggregation, (P3) label correction, (P4) robust loss, (P5) no supervision, (P6) data heterogeneity, and (P7) heterogeneous noise levels.}
\vspace{-5pt}
\begin{center}
\begin{tabular}{l|c|c|c|c|c|c|c}
	\toprule
	Method & P1 & P2 & P3 & P4 & P5 & P6 & P7 \\
	\midrule
	FOCUS\,\cite{focus} &  & \ding{51} &  &  & \ding{53} & $\triangle$ & $\triangle$ \\
	Tuor et al.\,\cite{tuor} & \ding{51} &  &  &  & \ding{53} & $\bigcirc$ & $\triangle$ \\
	RoFL\,\cite{rofl} & \ding{51} &  & \ding{51} &  & $\bigcirc$ & \ding{53} & $\triangle$ \\
	FedCorr\,\cite{fedcorr} &  &  & \ding{51} & \ding{51} & $\triangle$ & $\bigcirc$ & $\bigcirc$ \\
	LSR\,\cite{lsr} &  &  &  & \ding{51} & $\bigcirc$ & $\bigcirc$ & $\triangle$ \\
	\bottomrule
\end{tabular}
\vspace{-10pt}
\end{center}
\label{tab:previous_NFL}
\end{table}

%% file: tables/datasets.tex
\begin{table}[h] \footnotesize
\caption[Datasets and the common settings]{Datasets and the common settings. $\dagger$ indicates the pretrained architecture from ImageNet-1k.}
\addtolength{\tabcolsep}{-2pt}
\begin{center}
\begin{tabular}{lccc}
	\toprule
	Dataset & CIFAR-10 (CIFAR-10N) & CIFAR-100 & Clothing1M \\
	\midrule
	\# of train & 50,000 & 50,000 & 1,000,000 \\
	\# of test & 10,000 & 10,000 & 10,526 \\
	\# of classes & 10 & 100 & 14 \\
	\# of clients & 100 & 50 & 500 \\
	Participation ratio & 0.1 & 0.1 & 0.02 \\
	Model & ResNet-18 & ResNet-34 & ResNet-50$^\dagger$ \\
	Learning rate & 0.03 & 0.01 & 0.003 \\
	\bottomrule
\end{tabular}
\end{center}
\label{tab:datasets}
\vspace{-10pt}
\end{table}

%% file: tables/flr.tex
\begin{table*}[h]
\caption[Impelmentation details of FLR]{Impelmentation details of FLR for our experiments.}
\begin{center}
\begin{tabular}{lcc}
	\toprule
	Dataset & Case & $\lambda$  \\
	\midrule
	\multirow{2}{*}{CIFAR-10 (CIFAR-10N)}
    	& Symmetric noise & 2.0  \\
    	& Asymmetric noise & 3.0 \\
    	& Symmetric noise, w/ FedCorr & 3.0 \\
    	& Asymmetric noise, w/ FedCorr & 3.0   \\
    \midrule
    \multirow{2}{*}{CIFAR-100}
    	& Symmetric noise & 2.5   \\
    	& Asymmetric noise & 5.0  \\
    	& w/ Mixup & 2.0 \\
    	& w/ FedCorr & 5.0 \\
    \midrule
    \multirow{2}{*}{Clothing1M}
        & - & 2.0 \\
        & w/ FedCorr & 3.0  \\
	\bottomrule
\end{tabular}
\end{center}
\label{tab:flr}
\end{table*}

%% file: tables/er.tex
\begin{table}[h]
\caption[Ablation study on leLR]{Comparison with other label regularization methods, on CIFAR-10 with i.i.d. setting.}
\vspace{-5pt}
\begin{center}
\begin{tabular}{l|cc|cc}
	\toprule
	\multirow{4}{*}{Method} & \multicolumn{2}{c|}{Symmetric} & \multicolumn{2}{c}{Asymmetric} \\
	\cmidrule{2-5}
	& \multicolumn{2}{c|}{$\tau=0.0$} & \multicolumn{2}{c}{$\tau=0.4$} \\
	\cmidrule{2-5}
	& $\rho=0.6$ & $\rho=0.8$ & $\rho=0.6$ & $\rho=0.8$ \\
	\midrule
    	CE & 74.58$\pm$1.00 & 67.76$\pm$1.14 & 85.90$\pm$0.25 & 83.13$\pm$0.35 \\
    	EM & 86.14$\pm$0.45 & 82.90$\pm$1.01 & 88.25$\pm$0.20 & 86.17$\pm$0.69 \\
    	LS & 73.95$\pm$0.79 & 67.42$\pm$0.67 & 84.04$\pm$0.12 & 80.23$\pm$0.58 \\
    	NLS & 79.55$\pm$0.87 & 71.36$\pm$0.64 & 68.86$\pm$0.62 & 66.87$\pm$0.87 \\
    	TS & 82.28$\pm$0.65 & 76.38$\pm$1.34 & 87.45$\pm$0.34 & 84.97$\pm$0.63 \\
    	ER & \textbf{86.45$\pm$0.45} & \textbf{83.12$\pm$0.88} & \textbf{88.73$\pm$0.24} & \textbf{86.35$\pm$0.42} \\
	\bottomrule
\end{tabular}
\end{center}
\label{tab:er}
\vspace{-10pt}
\end{table}